\documentclass{article}

\usepackage{arxiv}

\usepackage[utf8]{inputenc} 
\usepackage[T1]{fontenc}    
\usepackage{hyperref}       
\usepackage{xurl}            
\usepackage{booktabs}       
\usepackage{amsfonts}       
\usepackage{amsmath,amssymb}
\usepackage{nicefrac}       
\usepackage{microtype}      
\usepackage{graphicx}
\usepackage{natbib}
\usepackage{doi}
\usepackage{multirow}
\usepackage{subcaption}
\usepackage{bm}
\usepackage{threeparttablex}
\usepackage{zi4}
\usepackage{pdfpages}

\title{HELM-BERT: A Transformer for Medium-sized Peptide Property Prediction}

\author{Seungeon Lee \\
	Graduate School of Medicine\\
	Kyoto University\\
	Kyoto, Japan \\
	\And
	Takuto Koyama \\
	Graduate School of Medicine\\
	Kyoto University\\
	Kyoto, Japan \\
	\And
	Itsuki Maeda \\
	Graduate School of Medicine\\
	Kyoto University\\
	Kyoto, Japan \\
	\And
	Shigeyuki Matsumoto\thanks{Corresponding author: matsumoto.shigeyuki.4z@kyoto-u.ac.jp} \\
	Graduate School of Medicine\\
	Kyoto University\\
	Kyoto, Japan \\
	\texttt{matsumoto.shigeyuki.4z@kyoto-u.ac.jp} \\
	\And
	Yasushi Okuno\thanks{Corresponding author: okuno.yasushi.4c@kyoto-u.ac.jp} \\
	Graduate School of Medicine\\
	Kyoto University\\
	Kyoto, Japan \\
	\texttt{okuno.yasushi.4c@kyoto-u.ac.jp} \\
}

\renewcommand{\shorttitle}{HELM-BERT}

\hypersetup{
pdftitle={HELM-BERT: A Transformer for Medium-sized Peptide Property Prediction},
pdfsubject={q-bio.BM, cs.LG},
pdfauthor={Seungeon Lee, Takuto Koyama, Itsuki Maeda, Shigeyuki Matsumoto, Yasushi Okuno},
pdfkeywords={HELM-BERT, HELM notation, cyclic peptide, membrane permeability, peptide-protein interaction, molecular representation},
}

\begin{document}
\maketitle

\begin{abstract}
Therapeutic peptides have emerged as a pivotal modality in modern drug discovery, occupying a chemically and topologically rich space. While accurate prediction of their physicochemical properties is essential for accelerating peptide development, existing molecular language models rely on representations that fail to capture this complexity. Atom-level SMILES notation generates long token sequences and obscures cyclic topology, whereas amino-acid-level representations cannot encode the diverse chemical modifications central to modern peptide design. To bridge this representational gap, the Hierarchical Editing Language for Macromolecules (HELM) offers a unified framework enabling precise description of both monomer composition and connectivity, making it a promising foundation for peptide language modeling. Here, we propose HELM-BERT, the first encoder-based peptide language model trained on HELM notation. Based on DeBERTa, HELM-BERT is specifically designed to capture hierarchical dependencies within HELM sequences. The model is pre-trained on a curated corpus of 39,079 chemically diverse peptides spanning linear and cyclic structures. HELM-BERT significantly outperforms state-of-the-art SMILES-based language models in downstream tasks, including cyclic peptide membrane permeability prediction and peptide--protein interaction prediction. These results demonstrate that HELM's explicit monomer- and topology-aware representations offer substantial data-efficiency advantages for modeling therapeutic peptides, bridging a long-standing gap between small-molecule and protein language models.
\end{abstract}

\keywords{HELM-BERT \and HELM notation \and cyclic peptide \and membrane permeability \and peptide--protein interaction \and molecular representation}

\section{Introduction}
Peptide therapeutics are an increasingly important drug modality, with more than eighty peptide drugs approved and over two hundred currently in clinical development \cite{zheng_therapeutic_2025}. Therapeutic peptides span a broad molecular-weight range (approximately 500--5,000~Da)~\cite{wang_therapeutic_2022} and bridge the gap between small molecules and biologics through their diverse chemical space and large interaction surfaces~\cite{wang_therapeutic_2022, vinogradov_macrocyclic_2019}. Their structural adaptability enables high-affinity engagement of large protein--protein interaction surfaces traditionally considered undruggable to small molecules, such as c-Myc and oncogenic KRAS~\cite{verdine_challenge_2007}, positioning peptides as an attractive modality for these challenging targets.

Developing peptide therapeutics requires satisfying a multiparametric objective profile. Beyond high affinity for the target, candidates must exhibit favorable physicochemical properties, such as metabolic stability and membrane permeability, to ensure efficacy in vivo. To achieve these properties, chemists routinely employ sophisticated strategies, including \textit{N}-methylation, amide-to-ester substitution, macrocyclization strategies, and incorporation of non-canonical residues, to rigidify the backbone and shield polar groups\cite{vinogradov_macrocyclic_2019}. However, as structural complexity increases, the empirical search for optimal candidates becomes a major bottleneck, as the synthesis and experimental characterization of such complex libraries are costly, labor-intensive, and time-consuming. This limitation motivates the use of computational methods capable of accurately predicting these critical properties to accelerate candidate screening and prioritize synthesis \cite{li_cycpeptmp_2024}.

Machine learning (ML) provides a powerful approach for predicting molecular properties. The accuracy of these models typically relies on the availability of large, high-quality training datasets. In this regard, pre-trained language models have emerged as a promising strategy, learning transferable representations from large unlabeled corpora \cite{ross_large-scale_2022}. Research in this field has primarily advanced along two distinct lines of molecular representation: For small molecules, atom-level models based on Simplified Molecular-Input Line-Entry System (SMILES)~\cite{weininger_smiles_1988} notation, such as ChemBERTa \cite{chithrananda_chemberta_2020} and MoLFormer--XL \cite{ross_large-scale_2022}, have achieved strong performance on standard benchmarks \cite{wu_moleculenet_2018}. For proteins, residue-level models, such as ESM-2 \cite{lin_evolutionary-scale_2023}, have enabled accurate prediction of structure and function. 

SMILES notation is designed to provide an unambiguous, atom-by-atom description of chemical structures~\cite{weininger_smiles_1988}. While effective for small molecules, applying this notation to peptides presents significant challenges. The large size of therapeutic peptides results in long token sequences that increase computational cost and dilutes local chemical information. Furthermore, SMILES encodes ring structures using implicit numerical identifiers that link non-adjacent atoms. This syntax creates non-local dependencies that obscure cyclic topology~\cite{wu_t-smiles_2024, jang_improving_2025}, making it difficult for sequence-based language models to reason about the complex molecular topology that governs essential physicochemical properties~\cite{ahlbach_beyond_2015, kelly_geometrically_2021}. Indeed, although recent efforts have extended SMILES-based pre-training to peptides, the notation remains cumbersome for large and highly complex peptides, prompting the exploration of alternative string representations~\cite{feller_peptide-aware_2025}.

Meanwhile, amino-acid-level representations operate at the residue level, offering a much more concise format than SMILES. However, applying standard protein language models to therapeutic peptides faces two fundamental limitations. First, their vocabulary is limited to the 20 canonical amino acids, precluding direct representation of diverse chemical modifications---such as D-amino acids, N-methylation, and non-canonical side chains---that are essential for peptide drug design. Second, the strictly linear sequence format cannot explicitly encode macrocyclic connectivity, failing to capture the constrained topologies that govern peptide stability and permeability. 

To address this representational gap between atomic resolution and residue-level abstraction, the Hierarchical Editing Language for Macromolecules (HELM) \cite{zhang_helm_2012} offers a compelling solution. HELM employs a hierarchical syntax that treats chemical monomers as fundamental units while explicitly defining connections and modifications, thereby bridging the gap between atomic precision and residue-level abstraction. This hybrid approach enables the precise description of non-canonical residues and macrocyclic topologies without generating the excessively long sequences inherent to SMILES. Despite these distinct advantages,  HELM's effectiveness in encoder-based models for property prediction remains unverified.

Here, we propose HELM-BERT, the first encoder-based language model trained on HELM notation. This model provides a unified, monomer-level, modification-aware representation for therapeutic peptides. To effectively capture both global topology and local chemical patterns in HELM sequences, we incorporate key architectural elements from DeBERTa \cite{he_deberta_2021}, including disentangled attention, Enhanced Mask Decoder (EMD), and n-gram induced encoding (nGiE). Pre-training of the model is performed using a curated corpus of 39,079 unique modified peptides spanning both linear and cyclic structures. The predictive performance is evaluated on two downstream tasks---membrane permeability and PPI prediction---showing that HELM-BERT significantly outperforms SMILES-based baselines on cyclic peptides and achieves competitive performance with large protein language models on natural-amino-acid peptides. Ablation studies identify disentangled attention as critical for learning effective representations from HELM notation, and embedding analysis reveals that HELM-BERT captures topological features more effectively than SMILES-based encoders.

By combining monomer-level representation with precise topological encoding, HELM-BERT establishes a robust framework for property prediction across diverse therapeutic peptides, both linear and cyclic, canonical and chemically modified. This unified framework holds the potential to accelerate screening and prioritize synthesis of structurally complex peptide candidates in modern drug design.

\section{Methods}
\subsection{Model Architecture}

\begin{figure*}[htbp]
    \centering
    \includegraphics[width=0.95\textwidth]{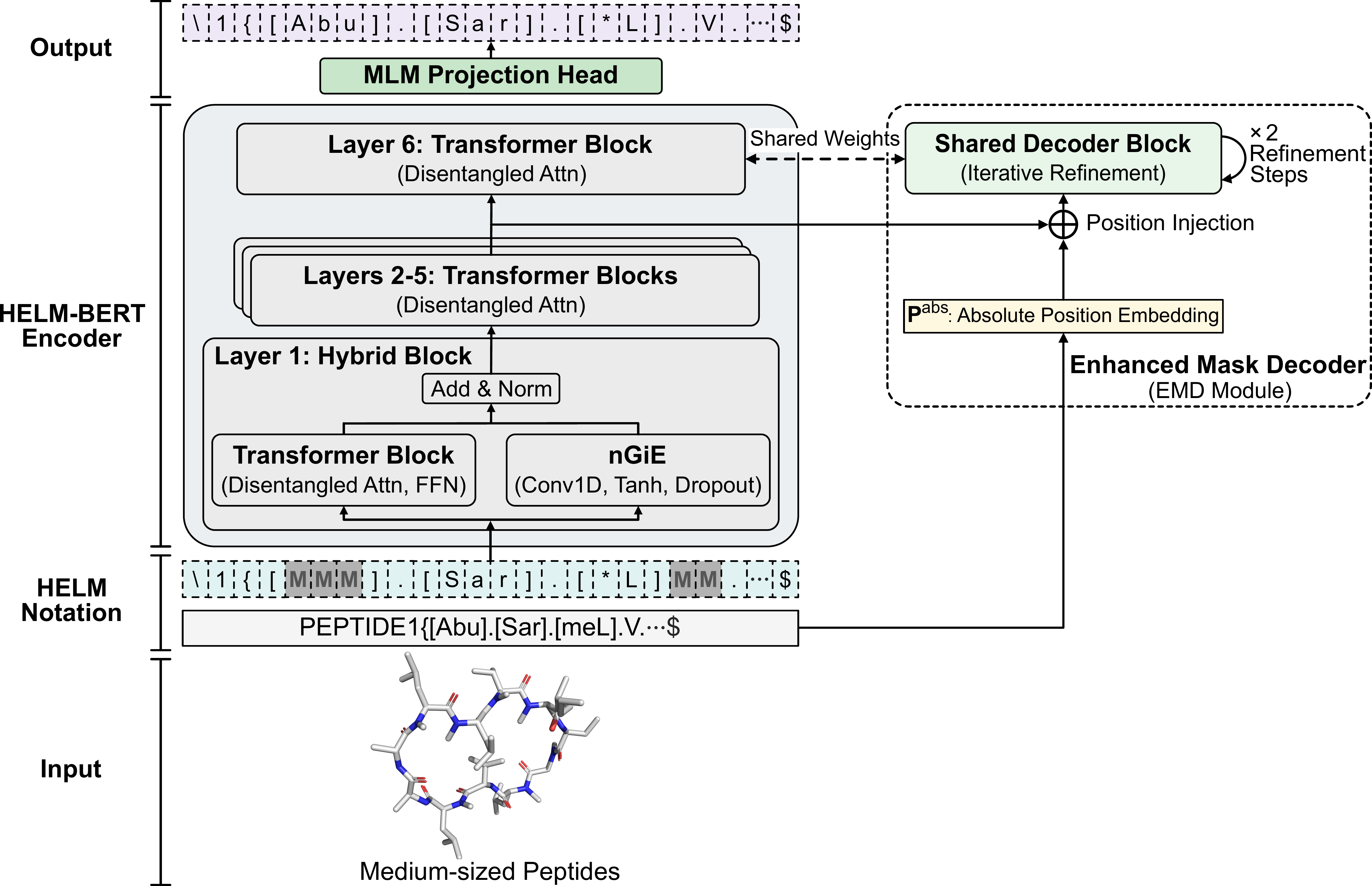}
    \caption{\textbf{Overview of HELM-BERT architecture.} Input peptides are converted to HELM notation, tokenized into monomer-level tokens, and subjected to span masking, where contiguous spans of tokens are masked (gray) during pre-training. The HELM-BERT encoder comprises a hybrid first layer combining disentangled self-attention with nGiE, followed by five transformer blocks with disentangled attention. The EMD receives the output of Layer~5, injects absolute position embeddings ($\mathbf{P}^{\text{abs}}$), and applies two weight-tied iterative refinement steps using the same parameters as Layer~6. The MLM projection head predicts the masked tokens (green).}
    \label{fig:architecture}
\end{figure*}

HELM-BERT is built upon the DeBERTa~\cite{he_deberta_2021} architecture (Figure~\ref{fig:architecture}). The backbone consists of a 6-layer transformer encoder with a hidden dimension of $H=768$ and $A=12$ attention heads. We first describe the tokenization scheme, then detail three architectural components adopted from DeBERTa: n-gram induced encoding (nGiE), disentangled attention, and Enhanced Mask Decoder (EMD).

\subsubsection{Tokenization and Input Representation}

To effectively process the hierarchical structure of macrocyclic peptides, we employed a dictionary-based character tokenizer with semantic compression, adopting the strategy of HELM-GPT~\cite{xu_helm-gpt_2024}, which is a generative model based on HELM notation. 
Unlike standard character-level tokenizers, our tokenizer explicitly encodes frequently occurring multi-character structural motifs as unique single-character markers, where common motifs such as \texttt{PEPTIDE} and \texttt{me} are mapped to dedicated markers (e.g., \texttt{/} and \texttt{*}, respectively) to preserve their semantic boundaries.

The vocabulary comprises 78 tokens including natural amino acids, structural delimiters, numbers, encoded polymer markers, and special tokens (\textbf{\texttt{[UNK]}}, \textbf{\texttt{[MASK]}}) for the Masked Language Modeling (MLM) objective.
Given an input HELM sequence of $n$ tokens, the model generates initial token embeddings $\mathbf{H}_0 \in \mathbb{R}^{n \times H}$.

\subsubsection{n-gram Induced Encoding}
HELM notation encodes recurrent chemical motifs, such as N-methylated residues, linker units, or side-chain protecting groups, as contiguous token sequences. To capture these local dependencies, we incorporate an nGiE layer following DeBERTa~\cite{he_deberta_2021}, implemented as a 1D convolution with kernel size $k=3$ applied in parallel to the first self-attention layer:

\begin{equation}
    \mathbf{H}_1^{\text{conv}} = \text{Tanh}(\text{Dropout}(\text{Conv1D}(\mathbf{H}_0, k=3)))
\end{equation}
\begin{equation}
    \mathbf{H}_1^{\text{attn}} = \operatorname{DSA}(\mathbf{H}_0)
\end{equation}
\begin{equation}
    \mathbf{H}_1 = \text{LayerNorm}(\mathbf{H}_1^{\text{attn}} + \mathbf{H}_1^{\text{conv}})
\end{equation}
where $\operatorname{DSA}$ denotes the Disentangled self-attention operation in the first Transformer encoder layer.

\subsubsection{Disentangled Attention Mechanism}
In HELM, macrocyclization and cross-links encoded in the connection table induce non-local couplings between distant positions in the linear monomer sequence. To model such distance-dependent interactions, we adopt the disentangled attention mechanism from DeBERTa~\cite{he_deberta_2021}, which decomposes attention scores into content–content and content–position terms:

\begin{equation}
    A_{i,j} = \underbrace{\mathbf{Q}_i^{\text{c}} (\mathbf{K}_j^{\text{c}})^\top}_{\text{(i) Content-to-Content}} 
    + \underbrace{\mathbf{Q}_i^{\text{c}} (\mathbf{K}_{\delta(i,j)}^{\text{r}})^\top}_{\text{(ii) Content-to-Position}} 
    + \underbrace{\mathbf{K}_j^{\text{c}} (\mathbf{Q}_{\delta(j,i)}^{\text{r}})^\top}_{\text{(iii) Position-to-Content}}
\end{equation}
Here, $\mathbf{Q}^{\text{c}}$ and $\mathbf{K}^{\text{c}}$ represent projected content vectors, while $\mathbf{Q}^{\text{r}}$ and $\mathbf{K}^{\text{r}}$ denote projected relative position vectors. $\delta(i,j)$ indicates the relative distance between token $i$ and $j$. The position-to-content term uses $\delta(j,i)$ following DeBERTa~\cite{he_deberta_2021}. We apply a scaling factor of $1/\sqrt{3d_h}$ instead of the standard $1/\sqrt{d_h}$ to account for the sum over the three components.

\subsubsection{Enhanced Mask Decoder}
This design allows the encoder to focus on learning rich relative-position patterns during pre-training, while absolute positions are provided as complementary information when disambiguating tokens with similar local context. Specifically, absolute position embeddings $\mathbf{P}^{\text{abs}}$ are withheld from the encoder and injected only into the query at the decoder stage. The EMD initializes its query at $t=0$ as:
\begin{equation}
    \mathbf{Q}^{(0)} = \mathbf{H}_{L-1} + \mathbf{P}^{\text{abs}}, \quad \mathbf{K} = \mathbf{V} = \mathbf{H}_{L-1}
\end{equation}
where $\mathbf{H}_{L-1}$ denotes the output of the penultimate encoder layer. The decoder then applies two iterative refinement steps ($t = 1, 2$), reusing the parameters of the final encoder layer (Layer~$L$):
\begin{equation}
    \mathbf{Q}^{(t)} = \text{TransformerBlock}_L(\mathbf{Q}^{(t-1)}, \mathbf{K}, \mathbf{V})
\end{equation}
The final output $\mathbf{Q}^{(2)}$ is passed to the MLM projection head, following DeBERTa~\cite{he_deberta_2021}.

\subsection{Pre-training}

\subsubsection{Data Sources}
We constructed a pre-training corpus from three public databases.

\noindent \textbf{ChEMBL v35}~\cite{mendez_chembl_2019}: A large-scale bioactivity database containing 22,045 entries with HELM notation, consisting of both linear and cyclic peptides with diverse chemical modifications including non-canonical amino acids and backbone modifications.

\noindent \textbf{CycPeptMPDB v1.2}~\cite{li_cycpeptmpdb_2023}: 8,466 cyclic peptides with experimental membrane permeability data ($\log P_{\text{app}}$) and HELM notation, representing the target domain for the downstream permeability prediction task.

\noindent \textbf{Propedia v2.3}~\cite{martins_propedia_2023}: 49,297 peptide--protein complex structures from the Protein Data Bank (PDB) with associated sequence and structural annotations.

\subsubsection{Data Processing}
\label{sec:data_processing}
For Propedia, we filtered out 18,113 entries (36.7\%) containing unknown residues (`X') or non-standard amino acids that could not be automatically converted to HELM notation, yielding 31,184 peptide--protein pairs. Peptide sequences were converted to HELM notation and SMILES using RDKit (version 2025.09.3)~\cite{landrum_rdkitrdkit_2025}.

Peptide deduplication was performed based on canonical SMILES, first within each dataset, then across datasets with priority ordering \textit{CycPeptMPDB} $>$ \textit{Propedia} $>$ \textit{ChEMBL}. The final pre-training corpus consists of \textbf{39,079 unique peptide sequences}: 21,879 from ChEMBL (56.0\%), 9,212 from Propedia (23.6\%), and 7,988 from CycPeptMPDB (20.4\%).

\subsubsection{Training Objective}
We pre-trained HELM-BERT using a Masked Language Modeling (MLM) objective with span masking~\cite{joshi_spanbert_2020, he_deberta_2021}. We masked 15\% of tokens in each sequence, with span lengths sampled from a geometric distribution ($p=0.2$) and clipped to the range $[1, 5]$. Masked spans were replaced following the standard 80–10–10 rule (80\% \texttt{[MASK]}, 10\% random token, 10\% unchanged).

The model was optimized using AdamW with a learning rate of $1 \times 10^{-4}$, weight decay of $0.01$, cosine annealing schedule, gradient clipping (max norm = 1.0), and a 32-bit floating point (FP32) precision. We trained the model with early stopping (patience $= 20$ epochs) and selected the checkpoint with the lowest validation loss.

\subsubsection{Embedding Quality Analysis}
To assess the information encoded in pre-trained representations, we conducted probing experiments on the pre-training corpus. For physicochemical property prediction, we used RDKit-computed LogP, molecular weight (MW), and topological polar surface area (TPSA) as regression targets. For structural feature classification, we used structure type (cyclic, lariat, linear) and number of rings as classification targets, both derived from HELM connectivity annotations. Structure type was categorized as cyclic (backbone-only cyclization via R1--R2 connections), lariat (side-chain involvement via R3), or linear (no intramolecular connections). Number of rings was defined as the count of intramolecular connection pairs in the HELM notation.

We evaluated representations using linear probing (5-fold cross-validation with L2-regularized linear models) and K-NN classification ($k=3$). Class separability was assessed using Silhouette score~\cite{rousseeuw_silhouettes_1987}, Davies-Bouldin index~\cite{davies_cluster_1979}, and Calinski-Harabasz index~\cite{calinski_dendrite_1974}. 
All evaluations used full-dimensional embeddings (768 dimensions for HELM-BERT and MoLFormer--XL, 768 dimensions for PeptideCLM). Statistical comparisons between models followed the procedure described in Section~\ref{sec:statistical_analysis}. K-NN and clustering metrics were computed as single-point estimates without cross-validation.

\subsection{Downstream Tasks}
\subsubsection{Membrane Permeability Prediction}
From the deduplicated CycPeptMPDB (7,988 entries), we removed 273 outliers with $\log P_{\text{app}} \le -10.0$ following the threshold used in prior work~\cite{li_cycpeptmpdb_2023}, yielding \textbf{7,715 samples}.
We employed 10-fold cross-validation. For each fold, 10\% of data was held out for testing, and the remaining 90\% was randomly split into training and validation sets (80\% and 10\% of total data, respectively). 

\subsubsection{Peptide--Protein Interaction Prediction}
From the filtered Propedia subset (Section~\ref{sec:data_processing}), duplicate peptide--receptor pairs were removed, yielding 20,057 unique pairs (9,212 peptides, 14,178 proteins, 9,634 PDB structures). Negative samples were generated by random pairing excluding known positives (1:4 positive-to-negative ratio).

For prediction, HELM-BERT encodes peptides (mean pooling, $H=768$) and ESM-2 (650M)~\cite{lin_evolutionary-scale_2023} encodes proteins (mean pooling, $H=1280$). Representations are concatenated and passed through a multi-layer perceptron (MLP). This dual-encoder design follows recent chemical genomics approaches that combine independently pretrained chemical and protein language models for interaction prediction, such as ChemGLaM \cite{koyama_chemglam_2024}.

We employed 5-fold cross-validation with two splitting strategies (hereafter referred to as \emph{Random Split} and \emph{Cluster-based Split}). For each fold, 20\% of data was held out for testing, and the remaining 80\% was randomly split into training and validation sets (70\% and 10\% of total data, respectively):
\begin{itemize}
    \item \textbf{Random Split}: Pair-grouped random splitting with no pair overlap across folds (20,057 positive pairs and 400,632 negative pairs).
    \item \textbf{Cluster-based Split}: K-means ($k=100$) clustering on atomic Cutoff Scanning Matrix (aCSM-ALL) signatures~\cite{pires_acsm_2013, martins_propedia_2023} (reduced from 3,600 to 50 dimensions via PCA), with clusters assigned to folds via constrained K-means ($k=5$, $\leq$15\% deviation). 
    Proteins appearing in multiple splits were assigned to their majority split, with ties resolved by prioritizing test over validation over training; pairs in non-assigned splits were removed to ensure no protein overlap within each fold (20,055 positive pairs and 394,337 negative pairs).
\end{itemize}

To address class imbalance, we used binary cross-entropy loss with a positive class weight of 4.0. 

\subsection{Experimental Setup}

\subsubsection{Baselines}

\paragraph{SMILES-based Models}
\begin{itemize}
    \item \textbf{MoLFormer--XL}~\cite{ross_large-scale_2022}: A 12-layer transformer encoder with 768 hidden dimensions, employing linear attention and rotary positional embeddings, pre-trained via masked language modeling on SMILES sequences from PubChem and ZINC. We used the publicly available checkpoint pre-trained on 10\% of the full dataset, as the complete model is not publicly released.
    \item \textbf{PeptideCLM}~\cite{feller_peptide-aware_2025}: A 6-layer RoFormer-based chemical language model with 768 hidden dimensions, pre-trained via masked language modeling on approximately 10 million modified peptides, 0.8 million natural peptides from SmProt, 10 million small molecules from PubChem, and 2.2 million patented molecules from SureChEMBL.
\end{itemize}

\paragraph{Sequence-based Models (PPI only)}
\begin{itemize}
    \item \textbf{ESM-2}~\cite{lin_evolutionary-scale_2023}: A transformer protein language model pre-trained via masked language modeling on UniRef protein sequences. We evaluated three variants with 35M, 150M, and 650M parameters.
    \item \textbf{Peptide Descriptors}~\cite{osorio_peptides_2015}: A feature extraction method that computes physicochemical descriptors from amino acid sequences using the \texttt{peptides} library in Python, including net charge, isoelectric point, hydrophobicity, hydrophobic moment, aliphatic index, and instability index.
\end{itemize}

\subsubsection{Evaluation Protocols}
We adopted three evaluation protocols to analyze the trade-off between representation quality and adaptability:
\begin{enumerate}
    \item \textbf{Full Fine-tuning}: End-to-end training of both the encoder and the task-specific head.
    \item \textbf{Head Fine-tuning}: Frozen encoder with a trainable non-linear prediction head (MLP).
    \item \textbf{Linear Probing}: Frozen encoder with a single linear layer.
\end{enumerate}

For membrane permeability prediction, models were evaluated under all three settings. For PPI, precomputed embeddings from frozen encoders were used with Linear Probing and Head Fine-tuning only.

\subsubsection{Implementation Details}
We use MLP heads as task-specific predictors in all downstream experiments, and specify their architecture for each task below. All downstream experiments used early stopping with patience 20 and maximum 200 epochs.

\paragraph{Membrane Permeability Prediction}
Table~\ref{tab:setup_permeability} summarizes the training configuration. All encoders are of comparable scale (43--54M parameters). For Fine-tuning settings, HELM-BERT uses a 3-layer MLP head with layer normalization, Gaussian Error Linear Unit (GELU) activation, and dropout ($p=0.1$) after each hidden layer (1.18M parameters). MoLFormer--XL uses its official 3-layer MLP head with GELU activation and dropout, without layer normalization (1.18M parameters). PeptideCLM uses its official 2-layer MLP head with Tanh activation (0.59M parameters). Linear Probing uses identical single linear layers across all models to isolate representation quality from head capacity.

\begin{table*}[htbp]
\centering
\begin{ThreePartTable}
\caption{Experimental configuration for membrane permeability prediction.}
\label{tab:setup_permeability}
\resizebox{\textwidth}{!}{%
\begin{tabular}{llcccl}
\toprule
\textbf{Model} & \textbf{Setting} & \textbf{Head Arch} & \textbf{Enc.\ Params} & \textbf{Head Params} & \textbf{Learning Rate} \\
\midrule
\multirow{3}{*}{HELM-BERT} 
    & Full FT & Residual MLP & \multirow{3}{*}{54.2M} & 1.18M & Enc: 3e-5 / Head: 1e-4 \\
    & Head FT & Residual MLP & & 1.18M & Head: 1e-4 \\
    & Linear  & Single Linear & & 0.77K & Head: 1e-3 \\
\midrule
\multirow{3}{*}{MoLFormer} 
    & Full FT & Official MLP & \multirow{3}{*}{44.4M} & 1.18M & Enc: 3e-5 / Head: 3e-5 \\
    & Head FT & Official MLP & & 1.18M & Head: 3e-5 \\
    & Linear  & Single Linear & & 0.77K & Head: 1e-3 \\
\midrule
\multirow{3}{*}{PeptideCLM} 
    & Full FT & Official MLP & \multirow{3}{*}{43.0M} & 0.59M & Enc: 5e-6 / Head: 5e-6 \\
    & Head FT & Official MLP & & 0.59M & Head: 5e-6 \\
    & Linear  & Single Linear & & 0.77K & Head: 1e-3 \\
\bottomrule
\end{tabular}%
}
\begin{tablenotes}
\scriptsize
\item All encoders are of comparable scale (43--54M parameters). HELM-BERT uses a custom 3-layer Residual MLP head; MoLFormer uses its official 3-layer MLP head (approximately 1.18M parameters each). PeptideCLM uses its official 2-layer MLP (0.59M parameters). Linear Probing uses identical single linear layers across all models. Optimizer: AdamW for HELM-BERT and Linear Probing; Adam for MoLFormer/PeptideCLM Fine-tuning.
\end{tablenotes}
\end{ThreePartTable}
\end{table*}

\paragraph{Peptide--Protein Interaction Prediction}
Table~\ref{tab:setup_ppi} summarizes the training configuration. All peptide encoders are paired with a frozen ESM-2 (650M) as the protein encoder. Peptide and protein representations are concatenated before prediction. To ensure fair comparison, we employed a unified 3-layer MLP head with residual connections for all Head Fine-tuning experiments, with a hidden dimension equal to the concatenated input dimension ($D_{pep} + D_{prot}$). Each of the two hidden layers consists of a linear transformation followed by GELU activation, layer normalization, dropout ($p=0.1$), and a residual connection. Linear Probing uses a single linear layer. This setup is intended to isolate the representational quality of the peptide encoders.\label{sec:methods_ppi}

\begin{table*}[htbp]
\centering
\begin{ThreePartTable}
\caption{Experimental configuration for PPI prediction.}
\label{tab:setup_ppi}
\resizebox{\textwidth}{!}{%
\begin{tabular}{llcccc}
\toprule
\textbf{Peptide Encoder} & \textbf{Setting} & \textbf{Pep.\ Enc.\ Params} & \textbf{Concat Dim} & \textbf{Head Params} & \textbf{Head LR} \\
\midrule
\multirow{2}{*}{HELM-BERT} 
    & Head FT & \multirow{2}{*}{54.2M} & \multirow{2}{*}{2048} & 8.4M & $1 \times 10^{-4}$ \\
    & Linear  & & & 2.0K & $1 \times 10^{-3}$ \\
\midrule
\multirow{2}{*}{MoLFormer-XL} 
    & Head FT & \multirow{2}{*}{44.4M} & \multirow{2}{*}{2048} & 8.4M & $1 \times 10^{-4}$ \\
    & Linear  & & & 2.0K & $1 \times 10^{-3}$ \\
\midrule
\multirow{2}{*}{PeptideCLM} 
    & Head FT & \multirow{2}{*}{43.0M} & \multirow{2}{*}{2048} & 8.4M & $1 \times 10^{-4}$ \\
    & Linear  & & & 2.0K & $1 \times 10^{-3}$ \\
\midrule
\multirow{2}{*}{ESM-2 (650M)} 
    & Head FT & \multirow{2}{*}{651M} & \multirow{2}{*}{2560} & 13.1M & $1 \times 10^{-4}$ \\
    & Linear  & & & 2.6K & $1 \times 10^{-3}$ \\
\midrule
\multirow{2}{*}{ESM-2 (150M)} 
    & Head FT & \multirow{2}{*}{148M} & \multirow{2}{*}{1920} & 7.4M & $1 \times 10^{-4}$ \\
    & Linear  & & & 1.9K & $1 \times 10^{-3}$ \\
\midrule
\multirow{2}{*}{ESM-2 (35M)} 
    & Head FT & \multirow{2}{*}{34M} & \multirow{2}{*}{1760} & 6.6M & $1 \times 10^{-4}$ \\
    & Linear  & & & 1.8K & $1 \times 10^{-3}$ \\
\midrule
\multirow{2}{*}{Peptide Descriptors} 
    & Head FT & \multirow{2}{*}{--} & \multirow{2}{*}{1382} & 3.8M & $1 \times 10^{-4}$ \\
    & Linear  & & & 1.4K & $1 \times 10^{-3}$ \\
\bottomrule
\end{tabular}%
}
\begin{tablenotes}
\scriptsize
\item All peptide encoders are paired with a frozen ESM-2 (650M) as the protein encoder. Peptide and protein representations are concatenated before prediction. Head Fine-tuning uses a unified 3-layer Residual MLP; Linear Probing uses a single linear layer. Head size scales with input dimension ($D_{pep} + D_{prot}$). Optimizer: AdamW.
\end{tablenotes}
\end{ThreePartTable}
\end{table*}

\paragraph{Ablation Studies}
To validate our architectural and pre-training choices, we conducted ablation experiments on the membrane permeability task under Full Fine-tuning. For architectural ablations, we compared HELM-BERT against variants that remove individual components:
\begin{itemize}
    \item \textbf{w/o Disentangled Attention}: replaces disentangled attention with standard self-attention, removing content-position decomposition.
    \item \textbf{w/o nGiE}: removes the convolutional n-gram encoding layer from the first Transformer block.
    \item \textbf{w/o EMD}: incorporates absolute position embeddings in the input layer instead of the decoder.
    \item \textbf{w/o Span Masking}: uses token-level MLM instead of span masking during pre-training.
    \item \textbf{Vanilla-BERT}: a standard 6-layer Transformer encoder with none of the above components (standard self-attention, no nGiE, input-layer position embeddings, token-level MLM).
\end{itemize}
For pre-training data ablations, we examined the contribution of each data source by removing one at a time (w/o ChEMBL, w/o Propedia, w/o CycPeptMPDB), and included a from-scratch baseline trained without any pre-training.

\subsubsection{Statistical Analysis}
\label{sec:statistical_analysis}
For all experiments, we report mean $\pm$ standard deviation over cross-validation folds.
For each task and metric, we compared the fold-wise test scores of HELM-BERT against those of each alternative model using the corrected resampled $t$-test for $k$-fold cross-validation~\cite{nadeau_inference_2003}.
To control the false discovery rate (FDR) across multiple comparisons, we applied the Benjamini--Hochberg procedure~\cite{benjamini_controlling_1995} with $q = 0.05$. All reported $p$-values are FDR-corrected unless otherwise noted. We also compute Cohen's $d$~\cite{cohen_statistical_2013} as an effect size for the fold-wise differences and refer to its magnitude in the text where relevant, using the conventional thresholds: $|d| < 0.2$ (negligible), $0.2 \leq |d| < 0.5$ (small), $0.5 \leq |d| < 0.8$ (medium), and $|d| \geq 0.8$ (large).

\section{Results and Discussion}
\subsection{Pre-training of HELM-BERT}
\begin{figure}[htbp]
    \centering
    \includegraphics[width=0.8\linewidth]{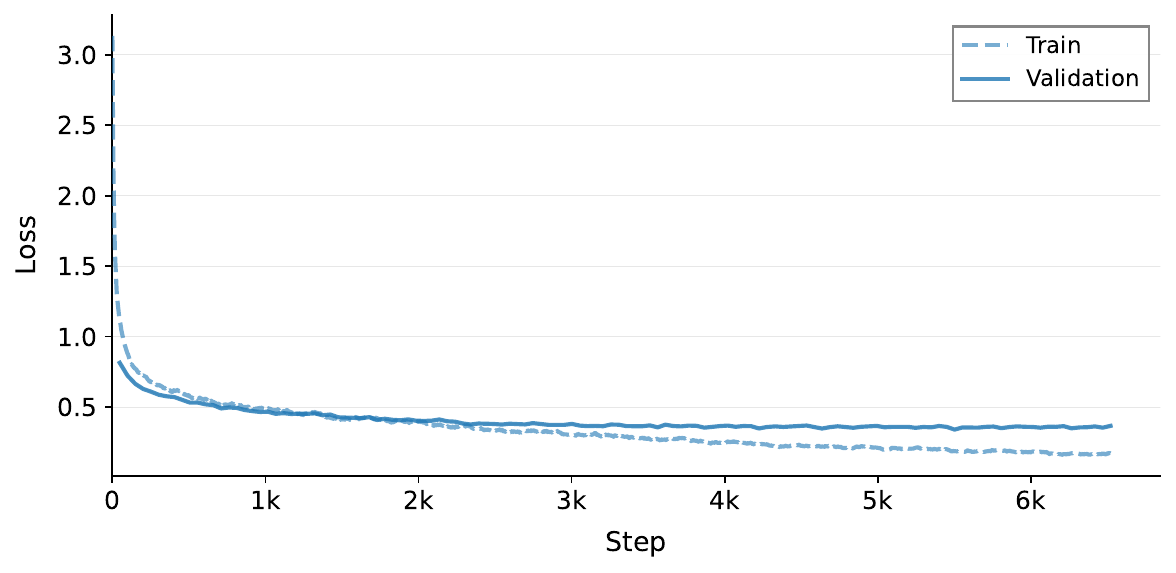}
    \caption{\textbf{Pre-training loss curves.} Training and validation MLM loss over the course of pre-training. The model was trained for 127 epochs with early stopping (patience = 20).}
    \label{fig:pretrain_loss}
\end{figure}

We pre-trained HELM-BERT on a curated corpus of 39,079 modified peptides compiled from ChEMBL, Propedia, and CycPeptMPDB, spanning diverse linear and cyclic structures (see Section~\ref{sec:data_processing} in Methods). Pre-training was performed on a single NVIDIA GH200 Grace Hopper Superchip using FP32 precision, requiring approximately 57~GB of GPU memory and 28~hours of training time. The model achieved the lowest validation loss (0.340) at epoch 107 (Figure~\ref{fig:pretrain_loss}). We selected this checkpoint as the pre-trained model and evaluated it on two downstream tasks that probe complementary aspects of peptide representation quality: membrane permeability prediction, which depends on backbone conformation and ring topology, and peptide--protein interaction prediction, which tests generalization to binding classification. 

\subsection{Membrane Permeability Prediction}

To assess the model's predictive performance on a critical property for therapeutic efficacy, we first focused on membrane permeability using the CycPeptMPDB benchmark. We conducted evaluations under three protocols that isolate different aspects of model performance: Full Fine-tuning (end-to-end training), Head Fine-tuning (frozen encoder with trainable MLP), and Linear Probing (frozen encoder with single linear layer). 

HELM-BERT achieved the highest performance across all settings (Table~\ref{tab:permeability}), with statistically significant improvements over both SMILES-based baselines. Under Full Fine-tuning, HELM-BERT achieved $R^2 = 0.717 \pm 0.035$, significantly exceeding MoLFormer--XL ($R^2 = 0.578 \pm 0.043$; $p < 0.001$, $d = 3.40$) and PeptideCLM ($R^2 = 0.536 \pm 0.025$; $p < 0.001$, $d = 7.18$). This advantage was consistent across Head Fine-tuning and Linear Probing, with all comparisons showing large effect sizes ($d > 2.3$, $p < 0.001$; Supplementary Tables~S4--S6). Linear Probing provides the most direct comparison of representation quality, as it uses identical linear layers across all models and eliminates confounding effects from differences in head architecture. In this setting, HELM-BERT maintained a substantial advantage over both SMILES-based baselines ($\Delta R^2 > 0.08$), suggesting that HELM notation encodes permeability-relevant structural features more effectively than atom-level SMILES representations. Notably, these improvements were achieved despite a substantially smaller pre-training corpus than SMILES-based models. We investigate the nature of these representations in Section~\ref{sec:embedding_analysis}. Fold-wise results are provided in Supplementary Table~S1; detailed statistical comparisons are reported in Supplementary Tables~S4--S6.

\begin{table*}[htbp]
\centering
\begin{ThreePartTable}
\caption{Performance comparison on the CycPeptMPDB permeability dataset.}
\label{tab:permeability}
\resizebox{\textwidth}{!}{%
\begin{tabular}{lrrrr}
\toprule
\textbf{Model} & \textbf{$R^2$ $\uparrow$} & \textbf{Pearson $r$ $\uparrow$} & \textbf{RMSE $\downarrow$} & \textbf{MAE $\downarrow$} \\
\midrule
\multicolumn{5}{c}{\textbf{Full Fine-tuning}} \\
\midrule
HELM-BERT            & $\mathbf{0.7172 \pm 0.0345}$ & $\mathbf{0.8493 \pm 0.0207}$ & $\mathbf{0.4164 \pm 0.0211}$ & $\mathbf{0.2946 \pm 0.0102}$ \\
MoLFormer--XL        & $\underline{0.5776 \pm 0.0434}^{\dagger}$ & $\underline{0.7673 \pm 0.0247}^{\dagger}$ & $\underline{0.5094 \pm 0.0267}^{\dagger}$ & $\underline{0.3668 \pm 0.0167}^{\dagger}$ \\
PeptideCLM           & $0.5360 \pm 0.0245^{\dagger}$ & $0.7413 \pm 0.0127^{\dagger}$ & $0.5344 \pm 0.0158^{\dagger}$ & $0.3847 \pm 0.0109^{\dagger}$ \\
\midrule
\multicolumn{5}{c}{\textbf{Head Fine-tuning}} \\
\midrule
HELM-BERT            & $\mathbf{0.6181 \pm 0.0343}$ & $\mathbf{0.7906 \pm 0.0199}$ & $\mathbf{0.4845 \pm 0.0231}$ & $\mathbf{0.3527 \pm 0.0143}$ \\
MoLFormer--XL        & $\underline{0.5510 \pm 0.0348}^{\dagger}$ & $\underline{0.7446 \pm 0.0218}^{\dagger}$ & $\underline{0.5255 \pm 0.0224}^{\dagger}$ & $\underline{0.3914 \pm 0.0130}^{\dagger}$ \\
PeptideCLM           & $0.4297 \pm 0.0256^{\dagger}$ & $0.6569 \pm 0.0193^{\dagger}$ & $0.5927 \pm 0.0245^{\dagger}$ & $0.4426 \pm 0.0141^{\dagger}$ \\
\midrule
\multicolumn{5}{c}{\textbf{Linear Probing}} \\
\midrule
HELM-BERT            & $\mathbf{0.4424 \pm 0.0293}$ & $\mathbf{0.6771 \pm 0.0136}$ & $\mathbf{0.5860 \pm 0.0243}$ & $\mathbf{0.4445 \pm 0.0221}$ \\
MoLFormer--XL        & $0.3070 \pm 0.0244^{\dagger}$ & $0.5611 \pm 0.0219^{\dagger}$ & $0.6535 \pm 0.0256^{\dagger}$ & $0.4950 \pm 0.0145^{\dagger}$ \\
PeptideCLM           & $\underline{0.3597 \pm 0.0213}^{\dagger}$ & $\underline{0.6035 \pm 0.0185}^{\dagger}$ & $\underline{0.6282 \pm 0.0246}^{\dagger}$ & $\underline{0.4703 \pm 0.0152}^{\dagger}$ \\
\bottomrule
\end{tabular}%
}
\begin{tablenotes}
\scriptsize
\item Best results are \textbf{bolded}, second-best are \underline{underlined}. $\uparrow$: higher is better, $\downarrow$: lower is better. $^{\dagger}$: significant difference from HELM-BERT (corrected resampled $t$-test with FDR correction, $p < 0.05$). Metrics: coefficient of determination ($R^2$), Pearson correlation ($r$), root mean squared error (RMSE), mean absolute error (MAE).
\end{tablenotes}
\end{ThreePartTable}
\end{table*}

\paragraph{Ablation Studies}
To identify which architectural components contribute most to learning HELM representations, we conducted ablation experiments on the membrane permeability task focusing on two key aspects: architectural components and pre-training corpus.

\begin{figure}[htbp]
    \centering
    \includegraphics[width=0.8\linewidth]{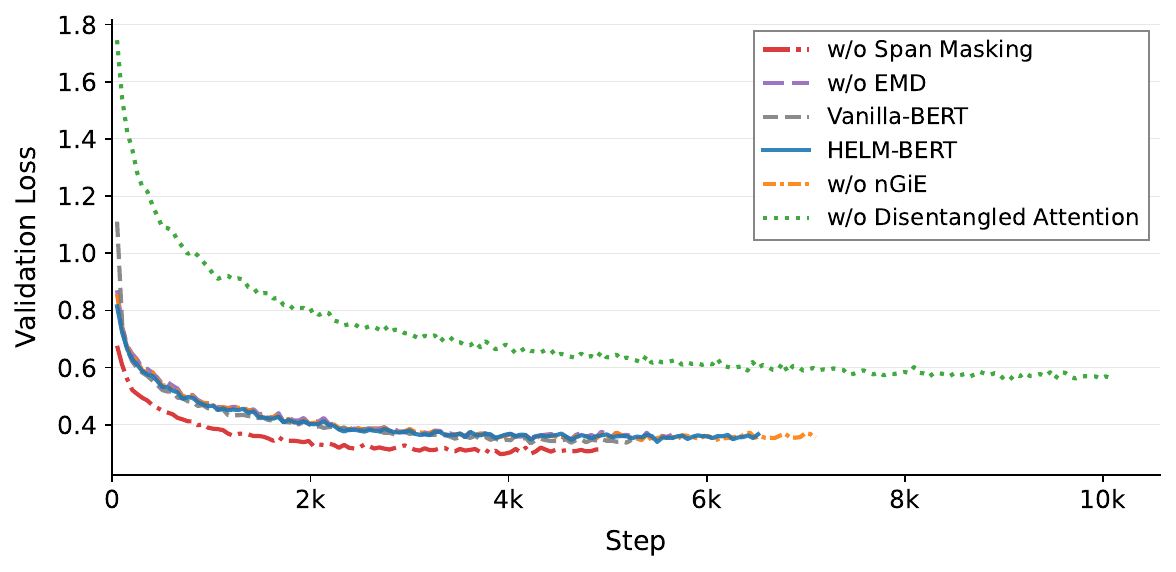}
    \caption{\textbf{Pre-training MLM loss curves under architectural ablations.} Validation loss for HELM-BERT and architectural variants over training epochs.}
    \label{fig:mlm_loss}
\end{figure}

Among the architectural components, removing disentangled attention produced the largest performance drop (Table~\ref{tab:ablation_architecture}): the gap between Vanilla-BERT and HELM-BERT ($\Delta R^2 = 0.065$) was largely explained by disentangled attention alone ($\Delta R^2 = 0.049$, representing 75\% of the total gap; $p = 0.001$, $d = 2.90$). This variant also led to destabilization of  pre-training (64\% more epochs, 70\% higher terminal loss; Figure~\ref{fig:mlm_loss}). To characterize how this ablation affects learned representations, we computed L2 norms of encoder weights across variants (Supplementary Table~S14). The variant without disentangled attention exhibited higher nGiE kernel norms ($37.6$ vs.\ $32.6$; +15\%) and position embedding norms ($44.7$ vs.\ $21.1$; +112\%), indicating compensatory reliance on absolute position information when relative position signals are unavailable. This indicates that disentangled attention is critical, accounting for the majority of the architectural contribution. Removing EMD, nGiE, or span masking individually showed no significant effects ($p > 0.28$, $d < 0.7$), suggesting complementary rather than essential contributions. Detailed statistical comparisons for architectural ablations are reported in Supplementary Table~S7.

For the pre-training corpus, the from-scratch baseline performed significantly worse ($R^2 = 0.664$; $p = 0.002$, $d = 2.48$), demonstrating that pre-training provides substantial benefit (Table \ref{tab:ablation_data}). We evaluated several data composition patterns and observed only minor performance variations, indicating that the specific data composition had limited impact. Intriguingly, excluding CycPeptMPDB, which contains cyclic peptides representing the target domain, showed no significant effect ($d = 0.35$). This finding indicates that HELM-BERT learns transferable representations rather than relying on task-specific structures. Detailed statistical comparisons for data ablations are reported in Supplementary Table~S8.

\begin{table*}[htbp]
\centering
\begin{ThreePartTable}
\caption{Ablation study on architecture and pre-training objective (CycPeptMPDB, Full Fine-tuning).}
\label{tab:ablation_architecture}
\resizebox{\textwidth}{!}{%
\begin{tabular}{lrrrr}
\toprule
\textbf{Variant} & \textbf{$R^2$ $\uparrow$} & \textbf{Pearson $r$ $\uparrow$} & \textbf{RMSE $\downarrow$} & \textbf{MAE $\downarrow$} \\
\midrule
HELM-BERT (full)              & $\mathbf{0.7172 \pm 0.0345}$ & $\mathbf{0.8493 \pm 0.0207}$ & $\mathbf{0.4164 \pm 0.0211}$ & $\mathbf{0.2946 \pm 0.0102}$ \\
w/o Disentangled Attention    & $0.6683 \pm 0.0303^{\dagger}$ & $0.8223 \pm 0.0184^{\dagger}$ & $0.4515 \pm 0.0202^{\dagger}$ & $0.3275 \pm 0.0134^{\dagger}$ \\
w/o EMD                       & $0.7045 \pm 0.0394$ & $0.8413 \pm 0.0224$ & $0.4256 \pm 0.0268$ & $0.3013 \pm 0.0152$ \\
w/o nGiE                      & $\underline{0.7129 \pm 0.0410}$ & $\underline{0.8462 \pm 0.0227}$ & $\underline{0.4192 \pm 0.0259}$ & $\underline{0.2973 \pm 0.0118}$ \\
w/o Span Masking              & $0.7064 \pm 0.0446$ & $0.8427 \pm 0.0260$ & $0.4239 \pm 0.0290$ & $0.3007 \pm 0.0172$ \\
Vanilla-BERT                  & $0.6523 \pm 0.0546^{\dagger}$ & $0.8114 \pm 0.0299^{\dagger}$ & $0.4616 \pm 0.0361^{\dagger}$ & $0.3289 \pm 0.0236^{\dagger}$ \\
\bottomrule
\end{tabular}%
}
\begin{tablenotes}
\scriptsize
\item All variants are pre-trained on the full corpus. ``Vanilla-BERT'' denotes a standard Transformer encoder without disentangled attention, nGiE, or EMD, trained with token-level MLM. Best results are \textbf{bolded}, second-best are \underline{underlined}. $\uparrow$: higher is better, $\downarrow$: lower is better. $^{\dagger}$ indicates significant difference from HELM-BERT (corrected resampled $t$-test with FDR correction, $p < 0.05$).
\end{tablenotes}
\end{ThreePartTable}
\end{table*}
\begin{table*}[htbp]
\centering
\begin{ThreePartTable}
\caption{Ablation study on pre-training data composition (CycPeptMPDB, Full Fine-tuning).}
\label{tab:ablation_data}
\resizebox{\textwidth}{!}{%
\begin{tabular}{lrrrr}
\toprule
\textbf{Variant} & \textbf{$R^2$ $\uparrow$} & \textbf{Pearson $r$ $\uparrow$} & \textbf{RMSE $\downarrow$} & \textbf{MAE $\downarrow$} \\
\midrule
HELM-BERT (full)              & $\underline{0.7172 \pm 0.0345}$ & $\underline{0.8493 \pm 0.0207}$ & $\underline{0.4164 \pm 0.0211}$ & $\underline{0.2946 \pm 0.0102}$ \\
w/o ChEMBL                    & $0.7024 \pm 0.0434$ & $0.8403 \pm 0.0246$ & $0.4268 \pm 0.0283$ & $0.3031 \pm 0.0192$ \\
w/o Propedia                  & $\mathbf{0.7259 \pm 0.0469}$ & $\mathbf{0.8542 \pm 0.0267}$ & $\mathbf{0.4091 \pm 0.0319}$ & $\mathbf{0.2922 \pm 0.0181}$ \\
w/o CycPeptMPDB               & $0.7094 \pm 0.0509$ & $0.8453 \pm 0.0281$ & $0.4213 \pm 0.0340$ & $0.2976 \pm 0.0207$ \\
From scratch                  & $0.6644 \pm 0.0411^{\dagger}$ & $0.8170 \pm 0.0246^{\dagger}$ & $0.4537 \pm 0.0275^{\dagger}$ & $0.3278 \pm 0.0133^{\dagger}$ \\
\bottomrule
\end{tabular}%
}
\begin{tablenotes}
\scriptsize
\item ``From scratch'' denotes a randomly initialized HELM-BERT encoder trained only on the downstream task. Best results are \textbf{bolded}, second-best are \underline{underlined}. $\uparrow$: higher is better, $\downarrow$: lower is better. $^{\dagger}$ indicates significant difference from HELM-BERT (corrected resampled $t$-test with FDR correction, $p < 0.05$).
\end{tablenotes}
\end{ThreePartTable}
\end{table*}

\subsection{Embedding Quality Analysis}
\label{sec:embedding_analysis}

To characterize the structural information encoded in pre-trained representations, we designed probing tasks targeting two complementary aspects of peptides: molecular properties (LogP, MW, TPSA) calculated directly from the atomic composition, and structural features (structure type, number of rings) that reflect macrocyclic topology (Table~\ref{tab:embedding_analysis}). We compared the performance of HELM-BERT against the two SMILES-based encoders, MoLFormer--XL and PeptideCLM, on these probing tasks using identical evaluation protocols.

Both HELM-BERT and SMILES-based encoders achieved high performance in predicting molecular properties calculated directly from the atomic composition ($R^2 > 0.95$), in contrast to the substantial performance differences observed in membrane permeability prediction. MoLFormer--XL significantly outperformed HELM-BERT on LogP ($p = 0.003$), while PeptideCLM showed no significant difference. For MW and TPSA, both SMILES-based encoders significantly outperformed HELM-BERT ($p < 0.001$, $d > 9$; Supplementary Table~S13). These results indicate that SMILES-based representations encode atomic-level molecular properties more effectively, likely because SMILES explicitly represents atom-level connectivity whereas HELM operates at the monomer level.

\begin{figure*}[htbp]
    \centering
    \begin{subfigure}[b]{0.32\textwidth}
        \centering
        \includegraphics[width=\textwidth]{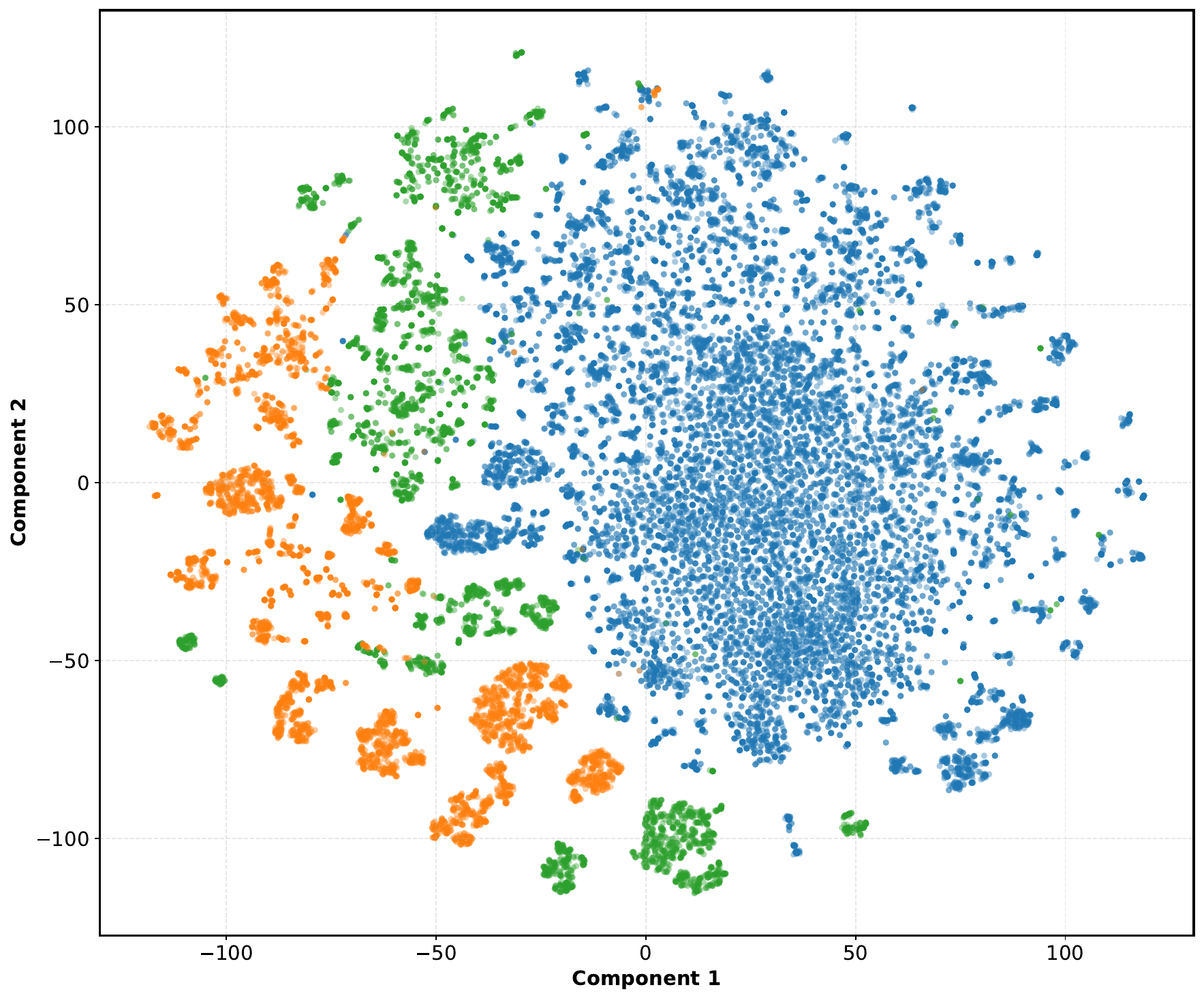}
        \caption{HELM-BERT}
    \end{subfigure}
    \hfill
    \begin{subfigure}[b]{0.32\textwidth}
        \centering
        \includegraphics[width=\textwidth]{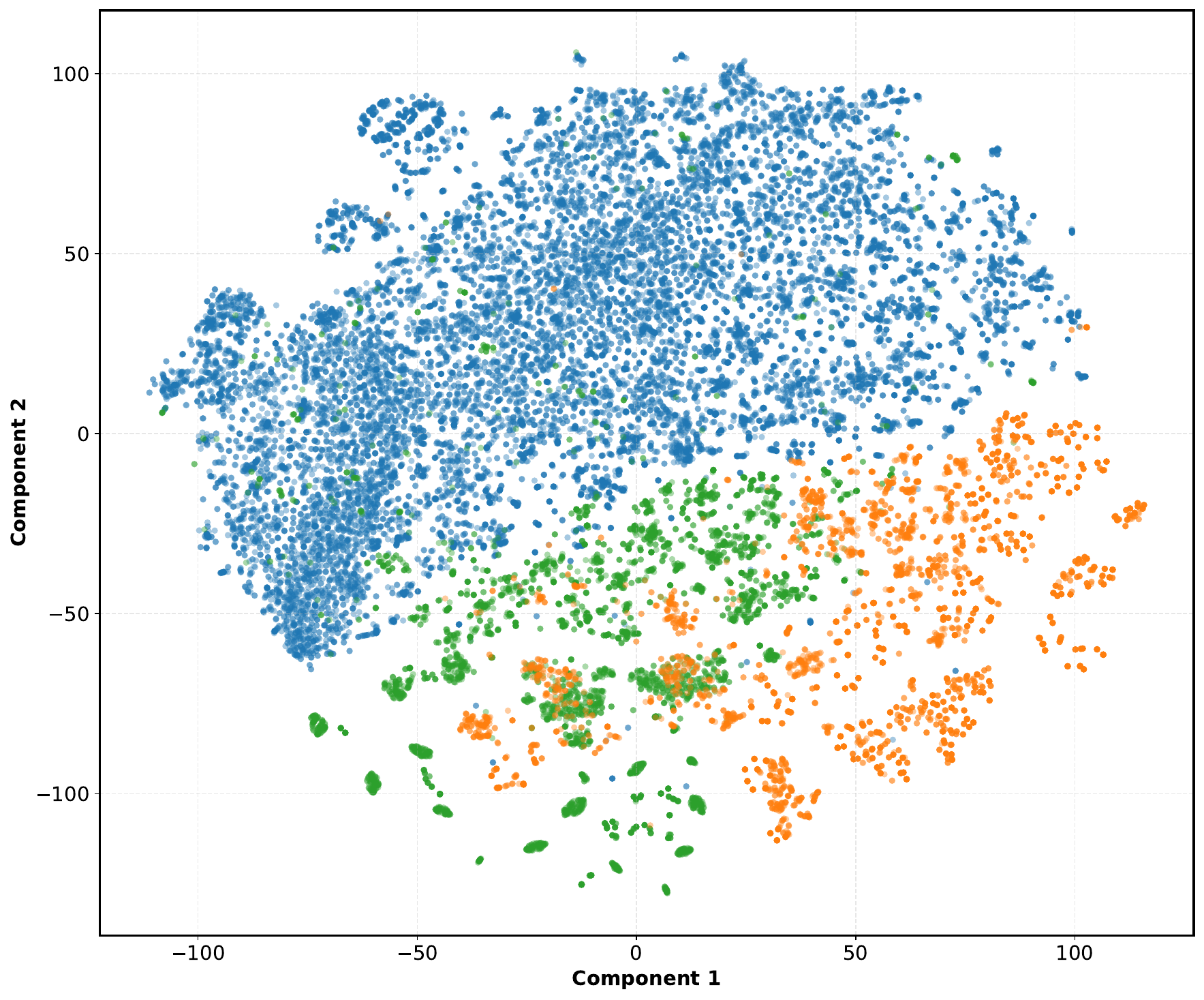}
        \caption{MoLFormer--XL}
    \end{subfigure}
    \hfill
    \begin{subfigure}[b]{0.32\textwidth}
        \centering
        \includegraphics[width=\textwidth]{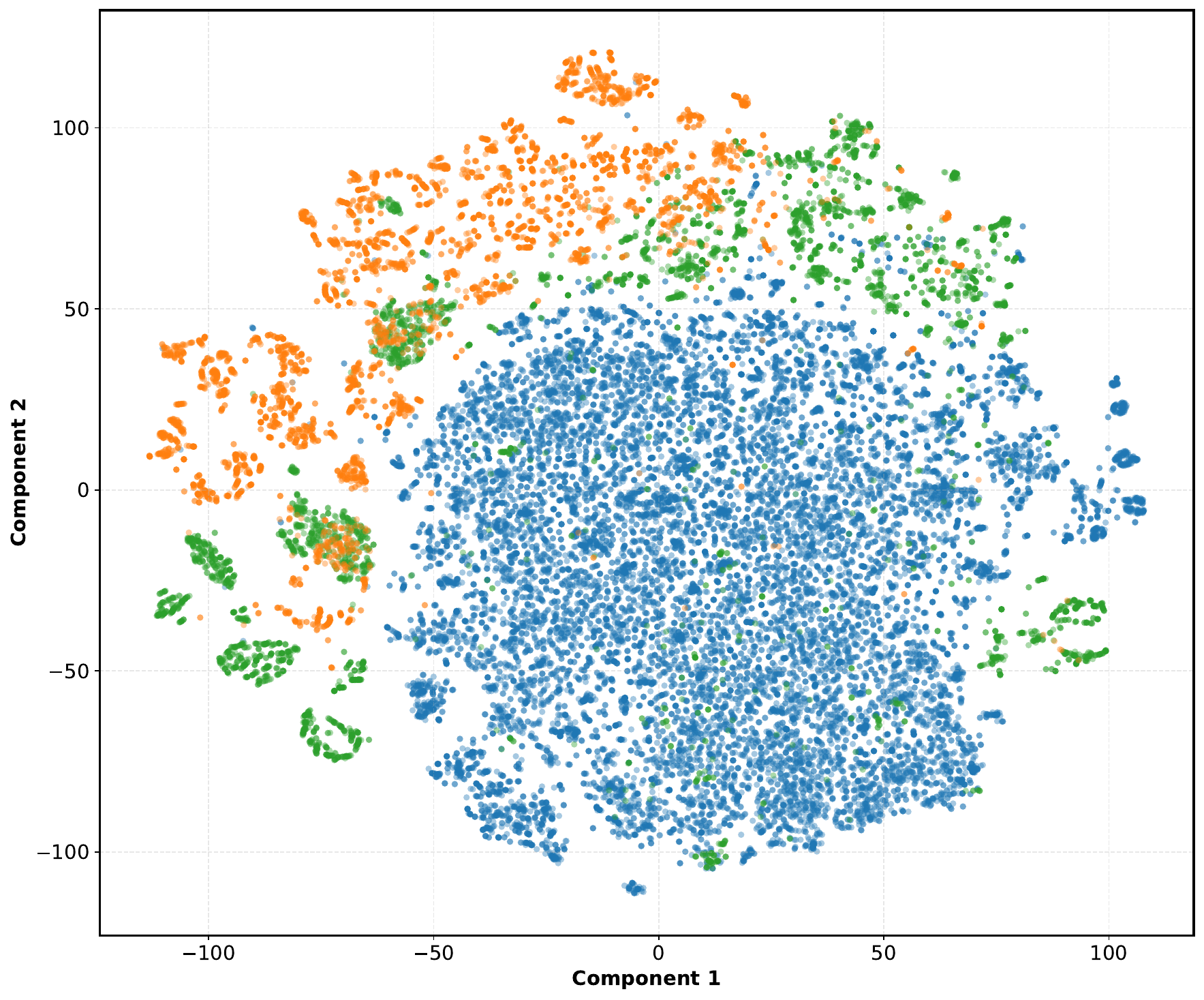}
        \caption{PeptideCLM}
    \end{subfigure}
    
    \vspace{0.3em}
    \includegraphics[width=0.6\textwidth]{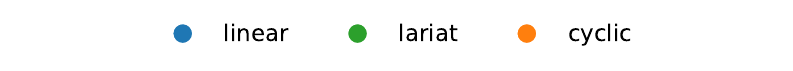}
    
    \caption{t-SNE projections of pre-trained embeddings colored by structure type.}
\label{fig:embedding_structure}
\end{figure*}

In contrast, HELM-BERT outperformed SMILES-based encoders on classification tasks of structural features that reflect macrocyclic topology. For Structure Type (cyclic, lariat, and linear peptides), HELM-BERT achieved $99.96 \pm 0.02$\% linear probing accuracy, significantly exceeding MoLFormer--XL ($98.10 \pm 0.15$\%) and PeptideCLM ($97.63 \pm 0.18$\%), with very large effect sizes ($d > 12$, $p < 0.001$ for both); similar patterns were observed for Number of Rings ($d > 10$, $p < 0.001$; Supplementary Table~S13). These findings were also supported by class separability metrics: HELM-BERT achieved the highest Silhouette score (0.106 vs.\ 0.072 for MoLFormer--XL and 0.096 for PeptideCLM), indicating greater within-class cohesion (Table~\ref{tab:embedding_analysis}). The t-SNE projections of the embeddings further illustrate this advantage, with HELM-BERT showing clearer separation between cyclic, lariat, and linear peptides (Figure~\ref{fig:embedding_structure} and Supplementary Figures~S2--S7). Low-dimensional (2D PCA) embedding analysis also showed consistent separation (Supplementary Table~S15). These results clearly demonstrate that HELM-BERT encodes discrete topological features more effectively than SMILES-based encoders.

Taken together, these results reveal a dichotomy: while SMILES-based encoders better capture atomic-level scalar properties, HELM-BERT more effectively encodes discrete topological features. Prior work has established that cyclic peptide permeability depends on backbone stereochemistry and N-methylation patterns, which determine conformation~\cite{ahlbach_beyond_2015}, as well as ring topology~\cite{kelly_geometrically_2021}. HELM-BERT's advantage in encoding topological features likely underlies its strong performance in membrane permeability prediction. This advantage may stem from HELM's explicit representation of topology: the linear monomer sequence and cyclization pattern are encoded separately, with ring closures and cross-links stored in an explicit connection list~\cite{zhang_helm_2012}.

\begin{table*}[htbp]
\centering
\begin{ThreePartTable}
\caption{Embedding quality evaluation of MLM-pre-trained encoders using full-dimensional representations.}
\label{tab:embedding_analysis}
\resizebox{\textwidth}{!}{%
\begin{tabular}{llccc}
\toprule
\textbf{Task} & \textbf{Metric} & \textbf{HELM-BERT} & \textbf{MoLFormer--XL} & \textbf{PeptideCLM} \\
\midrule
\multicolumn{5}{c}{\textbf{Physicochemical Properties (Regression)}} \\
\midrule
\multirow{2}{*}{LogP}
 & $R^2$ $\uparrow$  & $\underline{0.9535 \pm 0.0049}$ & $\mathbf{0.9638 \pm 0.0031}^{\dagger}$ & $0.9527 \pm 0.0018$ \\
 & MAE $\downarrow$  & $\underline{0.98 \pm 0.01}$ & $\mathbf{0.82 \pm 0.01}$ & $1.00 \pm 0.01$ \\
\midrule
\multirow{2}{*}{Molecular Weight}
 & $R^2$ $\uparrow$  & $0.9770 \pm 0.0003$ & $\underline{0.9842 \pm 0.0009}^{\dagger}$ & $\mathbf{0.9900 \pm 0.0001}^{\dagger}$ \\
 & MAE $\downarrow$  & $125.71 \pm 0.49$ & $\underline{96.25 \pm 1.19}$ & $\mathbf{82.28 \pm 1.19}$ \\
\midrule
\multirow{2}{*}{TPSA}
 & $R^2$ $\uparrow$  & $0.9779 \pm 0.0005$ & $\underline{0.9840 \pm 0.0008}^{\dagger}$ & $\mathbf{0.9880 \pm 0.0002}^{\dagger}$ \\
 & MAE $\downarrow$  & $55.03 \pm 0.45$ & $\underline{43.15 \pm 0.60}$ & $\mathbf{39.94 \pm 0.61}$ \\
\midrule
\multicolumn{5}{c}{\textbf{Structural Features (Classification \& Separability)}} \\
\midrule
\multirow{6}{*}{Structure Type}
 & Accuracy (K-NN) $\uparrow$   & $\mathbf{0.9993}$ & $\underline{0.9926}$ & $0.9910$ \\
 & Accuracy (Linear) $\uparrow$ & $\mathbf{0.9996 \pm 0.0002}$ & $\underline{0.9810 \pm 0.0015}^{\dagger}$ & $0.9763 \pm 0.0018^{\dagger}$ \\
 & MCC (Linear) $\uparrow$      & $\mathbf{0.9996}$ & $\underline{0.9714}$ & $0.9649$ \\
\cmidrule(lr){2-5}
 & Silhouette $\uparrow$        & $\mathbf{0.1060}$ & $0.0720$ & $\underline{0.0956}$ \\
 & Davies-Bouldin $\downarrow$  & $\mathbf{3.0179}$ & $\underline{3.1729}$ & $3.9966$ \\
 & Calinski-Harabasz $\uparrow$ & $\mathbf{2906}$ & $\underline{2686}$ & $2133$ \\
\midrule
\multirow{6}{*}{Number of Rings}
 & Accuracy (K-NN) $\uparrow$   & $\mathbf{0.9980}$ & $\underline{0.9923}$ & $0.9911$ \\
 & Accuracy (Linear) $\uparrow$ & $\mathbf{0.9975 \pm 0.0006}$ & $\underline{0.9788 \pm 0.0016}^{\dagger}$ & $0.9739 \pm 0.0027^{\dagger}$ \\
 & MCC (Linear) $\uparrow$      & $\mathbf{0.9979}$ & $\underline{0.9669}$ & $0.9603$ \\
\cmidrule(lr){2-5}
 & Silhouette $\uparrow$        & $\mathbf{0.0438}$ & $\underline{-0.0736}$ & $-0.0788$ \\
 & Davies-Bouldin $\downarrow$  & $\mathbf{2.2702}$ & $2.9537$ & $\underline{2.7056}$ \\
 & Calinski-Harabasz $\uparrow$ & $\mathbf{704}$ & $\underline{668}$ & $558$ \\
\bottomrule
\end{tabular}%
}
\begin{tablenotes}
\scriptsize
\item Linear probing and K-NN classification assess predictive performance; cluster validity indices (applied to ground-truth labels) quantify class separability. Best results are \textbf{bolded}, second-best are \underline{underlined}. $\uparrow$: higher is better, $\downarrow$: lower is better. $^{\dagger}$ indicates significant difference from HELM-BERT (corrected resampled $t$-test with FDR correction, $p < 0.05$); statistical tests were applied only to cross-validated metrics ($R^2$ and Linear Accuracy). For structural classification, all comparisons reached significance ($p < 0.001$, $|d| > 10$). For regression, MoLFormer--XL significantly outperformed HELM-BERT on MW ($p < 0.001$, $d = 10.83$), TPSA ($p < 0.001$, $d = 9.65$), and LogP ($p = 0.003$, $d = 4.52$); PeptideCLM significantly outperformed on MW ($p < 0.001$, $d = 33.92$) and TPSA ($p < 0.001$, $d = 14.79$), but not LogP ($p = 0.814$).
\end{tablenotes}
\end{ThreePartTable}
\end{table*}

\subsection{Peptide--Protein Interaction Prediction}
We evaluated peptide--protein interaction (PPI) prediction to test whether HELM-BERT's advantages extend beyond permeability prediction. Peptide encoders, such as HELM-BERT, MoLFormer--XL, and PeptideCLM, were paired with a frozen ESM-2 (650M) protein encoder, and their concatenated representations were passed to an MLP classifier (Supplementary Figure~S1). As the benchmark dataset constructed in this study comprises only peptides composed of natural amino acids, we also assessed simple Peptide Descriptors and ESM-2 variants, which are trained on millions of protein sequences, as peptide encoders. In this evaluation, we first evaluated models under Random Split, then examined generalization to unseen proteins using Cluster-based Split.

In the Random Split setting, HELM-BERT showed large effect sizes over SMILES-based encoders under Head Fine-tuning ($d = 2.5$--$3.6$;  Table~\ref{tab:ppi_random} and Supplementary Table~S9), though differences did not reach statistical significance after FDR correction. Under Linear Probing, HELM-BERT (ROC-AUC $= 0.612 \pm 0.005$) significantly outperformed SMILES-based encoders (MoLFormer--XL: $0.595 \pm 0.005$; PeptideCLM: $0.596 \pm 0.005$; $p < 0.01$, $d > 4$; Supplementary Table~S10). While most ESM-2 variants exhibited comparable performance to HELM-BERT, only ESM-2 (650M) achieved significantly higher linear separability ($p = 0.026$, $d = 3.13$). Fold-wise results are provided in Supplementary Table~S2; detailed statistical comparisons are reported in Supplementary Tables~S9--S10.

\begin{figure}[htbp]
    \centering
    \begin{subfigure}[b]{0.48\textwidth}
        \centering
        \includegraphics[width=\textwidth]{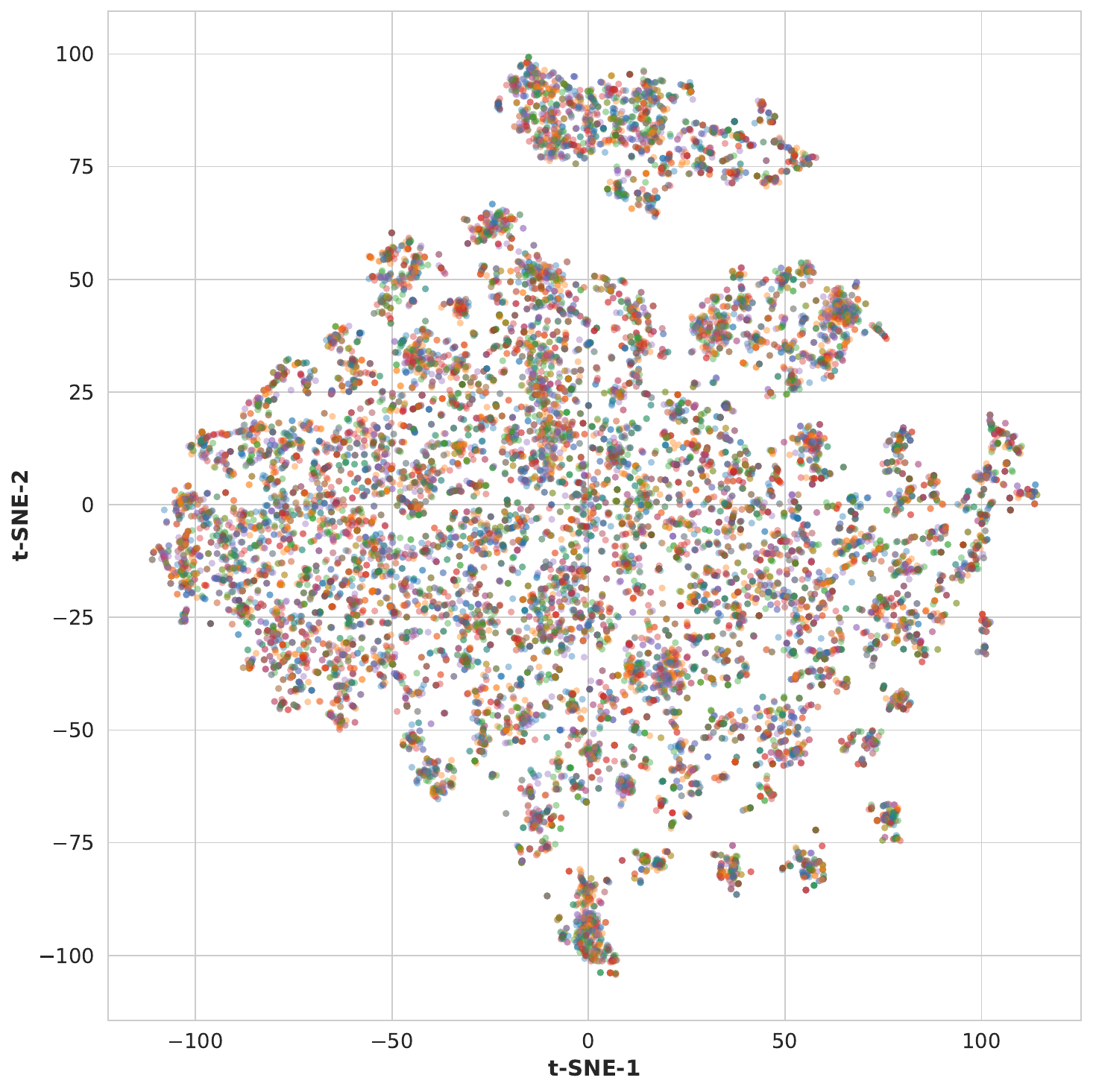}
        \caption{Random Split}
    \end{subfigure}
    \hfill
    \begin{subfigure}[b]{0.48\textwidth}
        \centering
        \includegraphics[width=\textwidth]{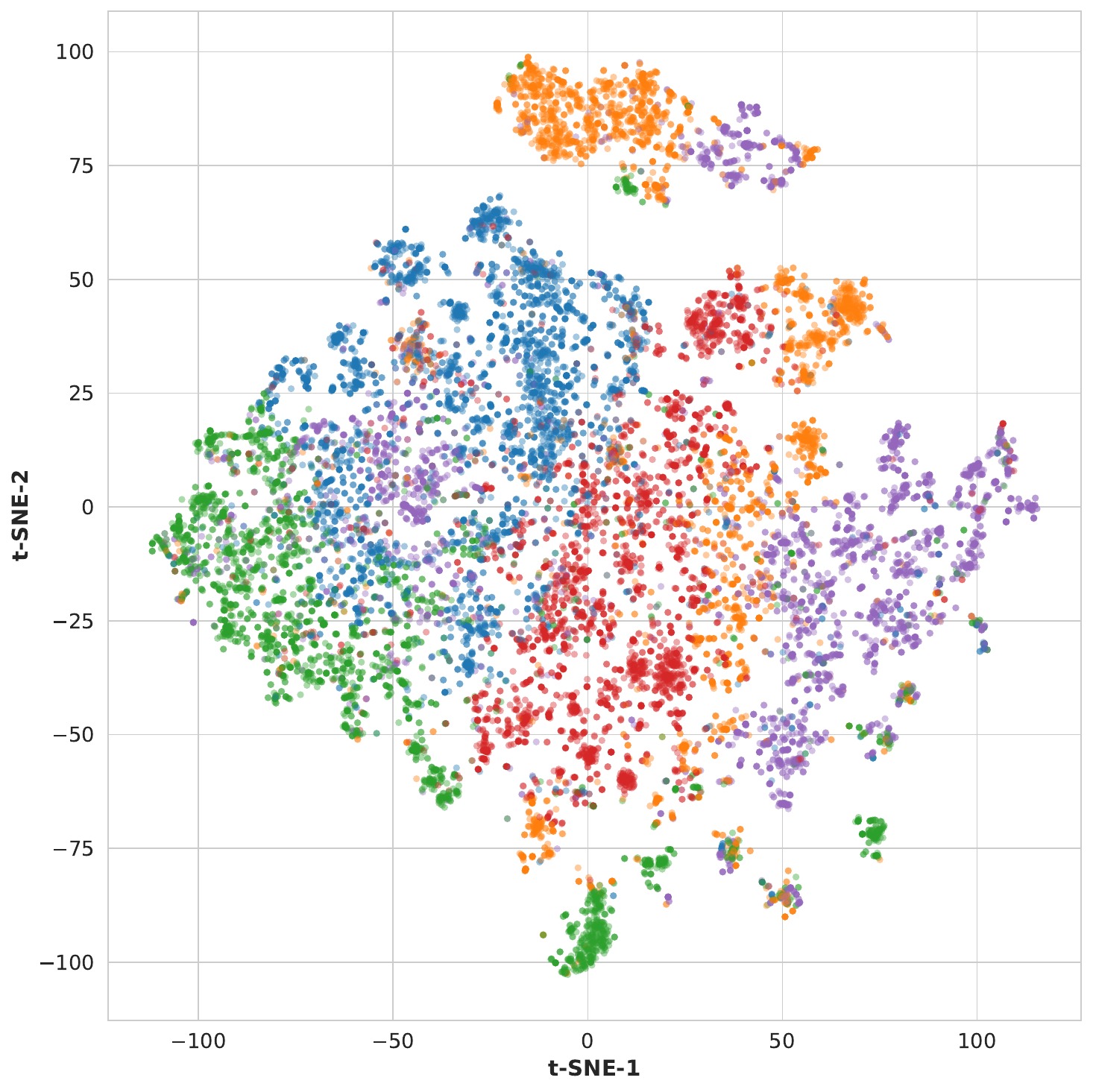}
        \caption{Cluster-based Split}
    \end{subfigure}
    
    \vspace{0.3em}
    \includegraphics[width=0.6\textwidth]{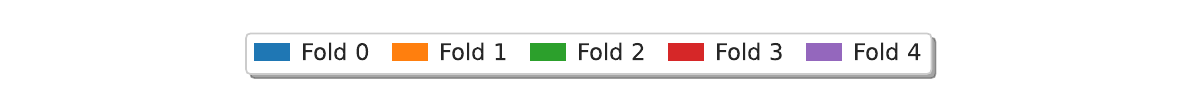}
    
    \caption{\textbf{t-SNE projections of PPI dataset splits in aCSM complex space.} Colors indicate fold assignment. The Cluster-based Split shows distinct spatial separation between folds, reflecting heterogeneous protein cluster distributions.}
    \label{fig:ppi_splits}
\end{figure}

To further examine model generalization to unseen proteins, we evaluated performance under the Cluster-based Split (Table~\ref{tab:ppi_acsm}), in which each fold tests on distinct complex clusters defined by aCSM signatures (Figure~\ref{fig:ppi_splits}). 

In this setting, HELM-BERT outperformed SMILES-based encoders under both Head Fine-tuning and Linear Probing: under Head Fine-tuning, HELM-BERT (ROC-AUC $= 0.771 \pm 0.042$) showed higher mean performance than MoLFormer--XL ($0.752 \pm 0.038$) and PeptideCLM ($0.742 \pm 0.026$), with medium-to-large effect sizes ($d = 0.5$--$0.8$; Supplementary Table~S11); under Linear Probing, HELM-BERT (ROC-AUC $= 0.566 \pm 0.022$) showed higher mean performance than MoLFormer--XL ($0.548 \pm 0.014$) and PeptideCLM ($0.542 \pm 0.011$), with large effect sizes ($d > 1.4$; Supplementary Table~S12). The predictive performance using ESM-2 variants was comparable to that of HELM-BERT (Head Fine-tuning: $0.771$ vs.\ $0.779$--$0.789$; Linear Probing: $0.566$ vs.\ $0.552$--$0.564$). The performance using Peptide Descriptors was consistently worse than that of ESM-2 variants, which have been shown to capture molecular structural features~\cite{lin_evolutionary-scale_2023}, suggesting the importance of utilizing topological features as observed in the embedding analysis of HELM-BERT. Fold-wise results are provided in Supplementary Table~S3; detailed statistical comparisons are reported in Supplementary Tables~S11--S12.

Consistent with the membrane permeability results, HELM-BERT outperformed SMILES-based encoders across both evaluation settings, indicating that HELM representations generalize effectively to diverse downstream tasks. Although, in the Cluster-based Split, pairwise comparisons did not reach statistical significance after FDR correction, the consistent effect sizes suggest practically meaningful advantages. Remarkably, HELM-BERT achieved comparable performance to ESM-2 variants despite being pre-trained on approximately 39,000 peptides, whereas ESM-2 was trained on millions of protein sequences. This result indicates favorable data efficiency of HELM-BERT. Furthermore, it should be emphasized that, unlike ESM-2, HELM-BERT can represent non-standard residues and chemical modifications, which may confer additional advantages for chemically modified peptides.

\begin{table*}[htbp]
\centering
\begin{ThreePartTable}
\caption{Performance comparison on Propedia PPI dataset with Random split.}
\label{tab:ppi_random}
\resizebox{\textwidth}{!}{%
\begin{tabular}{lrrrrr}
\toprule
\textbf{Model} & \textbf{Params} & \textbf{ROC-AUC} $\uparrow$ & \textbf{PR-AUC} $\uparrow$ & \textbf{MCC} $\uparrow$ & \textbf{Bal.\ Acc} $\uparrow$ \\
\midrule
\multicolumn{6}{c}{\textbf{MLP}} \\
\midrule
HELM-BERT            & 54.2M & $0.9420 \pm 0.0055$ & $0.8279 \pm 0.0181$ & $0.6705 \pm 0.0289$ & $0.8703 \pm 0.0080$ \\
ESM-2 (650M)         & 651M & $0.9416 \pm 0.0055$ & $0.8324 \pm 0.0152$ & $0.6704 \pm 0.0222$ & $0.8703 \pm 0.0089$ \\
ESM-2 (150M)         & 148M & $\underline{0.9434 \pm 0.0031}$ & $\underline{0.8418 \pm 0.0076}$ & $\underline{0.6822 \pm 0.0099}$ & $\underline{0.8740 \pm 0.0037}$ \\
ESM-2 (35M)          & 34M & $\mathbf{0.9466 \pm 0.0055}$ & $\mathbf{0.8474 \pm 0.0152}$ & $\mathbf{0.6886 \pm 0.0252}$ & $\mathbf{0.8787 \pm 0.0073}$ \\
PeptideCLM           & 43.0M & $0.9190 \pm 0.0078$ & $0.7807 \pm 0.0204$ & $0.6036 \pm 0.0331$ & $0.8405 \pm 0.0092$ \\
MoLFormer--XL        & 44.4M & $0.9218 \pm 0.0073$ & $0.7866 \pm 0.0170$ & $0.6178 \pm 0.0244$ & $0.8427 \pm 0.0134$ \\
Peptide Descriptors  & -- & $0.9320 \pm 0.0038$ & $0.8233 \pm 0.0090$ & $0.6589 \pm 0.0144$ & $0.8619 \pm 0.0061$ \\
\midrule
\multicolumn{6}{c}{\textbf{Linear}} \\
\midrule
HELM-BERT            & 54.2M & $0.6122 \pm 0.0049$ & $0.2704 \pm 0.0068$ & $\underline{0.1254 \pm 0.0077}$ & $\underline{0.5774 \pm 0.0053}$ \\
ESM-2 (650M)         & 651M & $\mathbf{0.6205 \pm 0.0026}^{\dagger}$ & $\mathbf{0.2706 \pm 0.0039}$ & $\mathbf{0.1315 \pm 0.0059}$ & $\mathbf{0.5809 \pm 0.0037}$ \\
ESM-2 (150M)         & 148M & $0.6109 \pm 0.0031$ & $0.2626 \pm 0.0042$ & $0.1235 \pm 0.0039$ & $0.5757 \pm 0.0026$ \\
ESM-2 (35M)          & 34M & $0.6047 \pm 0.0041$ & $0.2557 \pm 0.0053^{\dagger}$ & $0.1176 \pm 0.0066$ & $0.5719 \pm 0.0038$ \\
PeptideCLM           & 43.0M & $0.5956 \pm 0.0045^{\dagger}$ & $\underline{0.2625 \pm 0.0087}$ & $0.1066 \pm 0.0049^{\dagger}$ & $0.5658 \pm 0.0035$ \\
MoLFormer--XL        & 44.4M & $\underline{0.5949 \pm 0.0047}^{\dagger}$ & $0.2615 \pm 0.0023$ & $0.0991 \pm 0.0061^{\dagger}$ & $0.5609 \pm 0.0039^{\dagger}$ \\
Peptide Descriptors  & -- & $0.5584 \pm 0.0041^{\dagger}$ & $0.2311 \pm 0.0039^{\dagger}$ & $0.0665 \pm 0.0127^{\dagger}$ & $0.5412 \pm 0.0081^{\dagger}$ \\
\bottomrule
\end{tabular}%
}
\begin{tablenotes}
\scriptsize
\item Params indicates peptide encoder parameters. Best results are \textbf{bolded}, second-best are \underline{underlined}. $\uparrow$: higher is better. $^{\dagger}$: significant difference from HELM-BERT (corrected resampled $t$-test with FDR correction, $p < 0.05$). Metrics: area under the receiver operating characteristic curve (ROC-AUC), area under the precision--recall curve (PR-AUC), Matthews correlation coefficient (MCC)~\cite{matthews_comparison_1975, chicco_advantages_2020}, balanced accuracy.
\end{tablenotes}
\end{ThreePartTable}
\end{table*}
\begin{table*}[htbp]
\centering
\begin{ThreePartTable}
\caption{Performance comparison on Propedia PPI dataset with aCSM (cluster-based) split.}
\label{tab:ppi_acsm}
\resizebox{\textwidth}{!}{%
\begin{tabular}{lrrrrr}
\toprule
\textbf{Model} & \textbf{Params} & \textbf{ROC-AUC} $\uparrow$ & \textbf{PR-AUC} $\uparrow$ & \textbf{MCC} $\uparrow$ & \textbf{Bal.\ Acc} $\uparrow$ \\
\midrule
\multicolumn{6}{c}{\textbf{MLP}} \\
\midrule
HELM-BERT            & 54.2M & $0.7713 \pm 0.0420$ & $0.5090 \pm 0.0628$ & $0.3172 \pm 0.0737$ & $0.6873 \pm 0.0443$ \\
ESM-2 (650M)         & 651M & $\mathbf{0.7885 \pm 0.0339}$ & $\underline{0.5263 \pm 0.0397}$ & $0.3356 \pm 0.0545$ & $0.6989 \pm 0.0339$ \\
ESM-2 (150M)         & 148M & $0.7789 \pm 0.0473$ & $0.5118 \pm 0.0718$ & $\underline{0.3367 \pm 0.0775}$ & $0.6971 \pm 0.0426$ \\
ESM-2 (35M)          & 34M & $\underline{0.7882 \pm 0.0424}$ & $\mathbf{0.5282 \pm 0.0561}$ & $\mathbf{0.3464 \pm 0.0669}$ & $\mathbf{0.7046 \pm 0.0330}$ \\
PeptideCLM           & 43.0M & $0.7418 \pm 0.0264$ & $0.4516 \pm 0.0353$ & $0.2798 \pm 0.0377$ & $0.6712 \pm 0.0227$ \\
MoLFormer--XL        & 44.4M & $0.7516 \pm 0.0378$ & $0.4643 \pm 0.0496$ & $0.2938 \pm 0.0588$ & $\underline{0.6795 \pm 0.0330}$ \\
Peptide Descriptors  & -- & $0.7389 \pm 0.0572$ & $0.4627 \pm 0.0763$ & $0.2863 \pm 0.0762$ & $0.6700 \pm 0.0471$ \\
\midrule
\multicolumn{6}{c}{\textbf{Linear}} \\
\midrule
HELM-BERT            & 54.2M & $\mathbf{0.5656 \pm 0.0217}$ & $\mathbf{0.2333 \pm 0.0118}$ & $\mathbf{0.0685 \pm 0.0178}$ & $\mathbf{0.5414 \pm 0.0098}$ \\
ESM-2 (650M)         & 651M & $\underline{0.5644 \pm 0.0195}$ & $\underline{0.2276 \pm 0.0071}$ & $\underline{0.0647 \pm 0.0130}$ & $\underline{0.5373 \pm 0.0068}$ \\
ESM-2 (150M)         & 148M & $0.5587 \pm 0.0160$ & $0.2259 \pm 0.0107$ & $0.0590 \pm 0.0161$ & $0.5354 \pm 0.0086$ \\
ESM-2 (35M)          & 34M & $0.5517 \pm 0.0123$ & $0.2227 \pm 0.0099$ & $0.0520 \pm 0.0232$ & $0.5317 \pm 0.0148$ \\
PeptideCLM           & 43.0M & $0.5415 \pm 0.0106$ & $0.2234 \pm 0.0119$ & $0.0414 \pm 0.0139$ & $0.5250 \pm 0.0090$ \\
MoLFormer--XL        & 44.4M & $0.5484 \pm 0.0142$ & $0.2237 \pm 0.0087$ & $0.0599 \pm 0.0198$ & $0.5366 \pm 0.0123$ \\
Peptide Descriptors  & -- & $0.5310 \pm 0.0201$ & $0.2117 \pm 0.0093$ & $0.0274 \pm 0.0344$ & $0.5170 \pm 0.0210$ \\
\bottomrule
\end{tabular}%
}
\begin{tablenotes}
\scriptsize
\item Params indicates peptide encoder parameters. Best results are \textbf{bolded}, second-best are \underline{underlined}. $\uparrow$: higher is better. $^{\dagger}$: significant difference from HELM-BERT (corrected resampled $t$-test with FDR correction, $p < 0.05$).
\end{tablenotes}
\end{ThreePartTable}
\end{table*}

\clearpage
\section{Conclusion}

We present HELM-BERT, the first encoder-based language model trained on HELM notation for peptide property prediction. HELM-BERT bridges the representational gap between atom-level chemical language models and residue-level protein language models by providing a unified framework that captures both monomer-level chemistry and macrocyclic topology.

On cyclic peptide permeability prediction, HELM-BERT significantly outperforms SMILES-based baselines. Ablation studies identify disentangled attention as critical for learning effective representations from HELM notation. Embedding analysis reveals that HELM-BERT captures topological features more effectively than SMILES-based encoders, which may underlie its strong performance on topology-dependent properties. In peptide--protein interaction tasks, HELM-BERT achieves competitive performance with protein language models despite training on a corpus orders of magnitude smaller.

These results establish the effectiveness of hierarchical, topology-aware representations for therapeutic peptides. However, a fundamental challenge remains: HELM-BERT requires HELM annotations that are either natively available or obtainable via automated conversion, which limits both the scale of pre-training corpora and the scope of evaluation. Addressing this challenge will require community efforts toward HELM standardization, which would enable larger and more diverse pre-training corpora. 

As HELM adoption grows, future work can explore architectures that more explicitly model HELM's compositional semantics---for example, graph-based modules that jointly encode monomer sequences and connectivity graphs. Such advances would further close the gap between small-molecule and protein representations, accelerating the design of structurally complex therapeutic peptides.

\section*{Supplementary Information}
Supporting Information includes: (S1) PPI prediction framework; (S2) Fold-wise performance results; (S3) Statistical comparison of models; (S4) Component activation analysis; (S5) Embedding visualizations colored by structure type, source dataset, and physicochemical properties; (S6) Low-dimensional embedding analysis.

\section*{Acknowledgements}
Computational resources were provided by the Miyabi supercomputer (JCAHPC). This research was supported by Japan Agency for Medical Research and Development (AMED) under Grant Number JP25nk0101112.

\section*{Author Contributions}
T.K. and S.M. conceptualized the study. S.L. developed the methodology, implemented the software, performed the experiments, and analyzed the data. S.L. and T.K. were responsible for data curation. S.L. wrote the original draft of the manuscript. T.K., I.M., S.M., and Y.O. provided supervision and reviewed and edited the manuscript. Y.O. acquired funding. All authors have read and agreed to the published version of the manuscript.

\section*{Competing Interests}
The authors declare no competing interests.

\section*{Data Availability}
The HELM-BERT model, training code, and pre-trained weights are available at \url{https://github.com/clinfo/HELM-BERT.git}. The pretrained checkpoint is publicly available at \url{https://drive.google.com/drive/folders/1XKtmlSvFwl3smxVqyOfnSKi8lFe54Pyi?usp=sharing}.

\section*{Code Availability}
Available at \url{https://github.com/clinfo/HELM-BERT.git}.

\bibliographystyle{unsrtnat}
\bibliography{references}  






\clearpage


\section*{Supplementary material}

# Supplementary Materials for "HELM-BERT: A Transformer for Medium-sized Peptide Property Prediction"

Seungeon Lee[1], Takuto Koyama[1], Itsuki Maeda[1], Shigeyuki Matsumoto[1*], Yasushi Okuno[1*]

[1]Graduate School of Medicine, Kyoto University, Kyoto, Japan.

[*]Corresponding author(s). E-mail(s):
matsumoto.shigeyuki.4z@kyoto-u.ac.jp; okuno.yasushi.4c@kyoto-u.ac.jp

## S1. PPI Prediction Framework

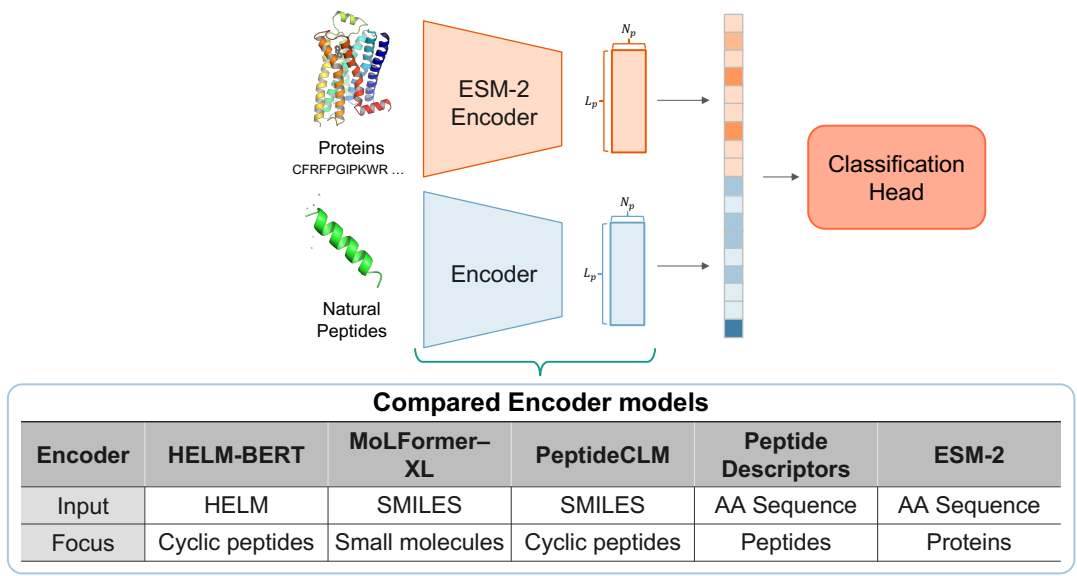

Figure S1: **Dual-encoder framework for PPI prediction.** Protein sequences are encoded by a frozen ESM-2 (650M), while peptide sequences are encoded by one of the compared encoders. Representations are concatenated and passed to an MLP classification head. The table summarizes the input format and primary training focus of each peptide encoder.

# S2. Fold-wise Performance Results

To support the reliability of our cross-validation results, we report fold-wise performance for the main downstream tasks. These tables demonstrate that performance variations across folds are consistent across models, and that no single fold disproportionately affects the averaged results.

## S2.1 CycPeptMPDB permeability prediction

Table S1 reports fold-wise performance for HELM-BERT, MoLFormer–XL, and PeptideCLM on CycPeptMPDB permeability prediction across all three evaluation protocols. HELM-BERT consistently outperforms the baselines across all 10 folds, demonstrating robust performance regardless of the data partition.

Table S1: Fold-wise performance on CycPeptMPDB permeability prediction across all evaluation protocols.

| Fold | $R^2$ ↑ | | | Pearson $r$ ↑ | | | RMSE ↓ | | | MAE ↓ | | |
|---|---|---|---|---|---|---|---|---|---|---|---|---|
| | HB | MF | PC | HB | MF | PC | HB | MF | PC | HB | MF | PC |
| | | | | | | Full Fine-tuning | | | | | | |
| 0 | **0.7298** | 0.5909 | 0.5159 | **0.8553** | 0.7816 | 0.7357 | **0.4221** | 0.5194 | 0.5650 | **0.2953** | 0.3620 | 0.3927 |
| 1 | **0.7310** | 0.5913 | 0.5542 | **0.8565** | 0.7717 | 0.7511 | **0.4324** | 0.5329 | 0.5567 | **0.3081** | 0.3860 | 0.3906 |
| 2 | **0.7318** | 0.6424 | 0.5216 | **0.8604** | 0.8051 | 0.7245 | **0.3901** | 0.4505 | 0.5211 | **0.2815** | 0.3321 | 0.3846 |
| 3 | **0.6836** | 0.5717 | 0.5299 | **0.8296** | 0.7619 | 0.7453 | **0.4307** | 0.5011 | 0.5250 | **0.2981** | 0.3688 | 0.3790 |
| 4 | **0.7064** | 0.5154 | 0.5222 | **0.8429** | 0.7332 | 0.7310 | **0.4197** | 0.5391 | 0.5353 | **0.3021** | 0.3831 | 0.3970 |
| 5 | **0.7684** | 0.5643 | 0.5837 | **0.8820** | 0.7542 | 0.7662 | **0.3908** | 0.5361 | 0.5240 | **0.2852** | 0.3829 | 0.3682 |
| 6 | **0.7702** | 0.6325 | 0.5628 | **0.8790** | 0.7967 | 0.7510 | **0.3857** | 0.4878 | 0.5320 | **0.2805** | 0.3590 | 0.3883 |
| 7 | **0.6815** | 0.5922 | 0.5454 | **0.8264** | 0.7761 | 0.7454 | **0.4417** | 0.4998 | 0.5277 | **0.3067** | 0.3742 | 0.3855 |
| 8 | **0.6887** | 0.5067 | 0.5084 | **0.8340** | 0.7282 | 0.7317 | **0.4104** | 0.5166 | 0.5158 | **0.2877** | 0.3692 | 0.3652 |
| 9 | **0.6801** | 0.5686 | 0.5162 | **0.8265** | 0.7642 | 0.7309 | **0.4400** | 0.5109 | 0.5411 | **0.3002** | 0.3504 | 0.3955 |
| avg | **0.7172** | 0.5776 | 0.5360 | **0.8493** | 0.7673 | 0.7413 | **0.4164** | 0.5094 | 0.5344 | **0.2946** | 0.3668 | 0.3847 |
| std | 0.0345 | 0.0434 | 0.0245 | 0.0207 | 0.0247 | 0.0127 | 0.0211 | 0.0267 | 0.0158 | 0.0102 | 0.0167 | 0.0109 |
| | | | | | | Head Fine-tuning | | | | | | |
| 0 | **0.6035** | 0.5245 | 0.4466 | **0.7846** | 0.7263 | 0.6693 | **0.5114** | 0.5600 | 0.6041 | **0.3622** | 0.4120 | 0.4460 |
| 1 | **0.6532** | 0.5598 | 0.4069 | **0.8113** | 0.7489 | 0.6406 | **0.4909** | 0.5532 | 0.6421 | **0.3568** | 0.4052 | 0.4659 |
| 2 | **0.6615** | 0.5768 | 0.4243 | **0.8135** | 0.7623 | 0.6519 | **0.4383** | 0.4901 | 0.5716 | **0.3279** | 0.3731 | 0.4288 |
| 3 | **0.5998** | 0.5468 | 0.4370 | **0.7836** | 0.7407 | 0.6617 | **0.4844** | 0.5154 | 0.5745 | **0.3480** | 0.3834 | 0.4270 |
| 4 | **0.6224** | 0.5458 | 0.4651 | **0.7966** | 0.7423 | 0.6840 | **0.4759** | 0.5220 | 0.5664 | **0.3524** | 0.3900 | 0.4270 |
| 5 | **0.6213** | 0.5834 | 0.4408 | **0.7912** | 0.7648 | 0.6648 | **0.4997** | 0.5242 | 0.6073 | **0.3730** | 0.3912 | 0.4500 |
| 6 | **0.6727** | 0.6168 | 0.4442 | **0.8204** | 0.7861 | 0.6686 | **0.4603** | 0.4980 | 0.5998 | **0.3411** | 0.3771 | 0.4527 |
| 7 | **0.5814** | 0.5436 | 0.3866 | **0.7638** | 0.7388 | 0.6232 | **0.5064** | 0.5288 | 0.6130 | **0.3687** | 0.4014 | 0.4589 |
| 8 | **0.5855** | 0.5034 | 0.3960 | **0.7772** | 0.7222 | 0.6329 | **0.4736** | 0.5184 | 0.5717 | **0.3369** | 0.3803 | 0.4322 |
| 9 | **0.5794** | 0.5095 | 0.4501 | **0.7641** | 0.7140 | 0.6720 | **0.5045** | 0.5448 | 0.5769 | **0.3595** | 0.4000 | 0.4370 |
| avg | **0.6181** | 0.5510 | 0.4297 | **0.7906** | 0.7446 | 0.6569 | **0.4845** | 0.5255 | 0.5927 | **0.3527** | 0.3914 | 0.4426 |
| std | 0.0343 | 0.0348 | 0.0256 | 0.0199 | 0.0218 | 0.0193 | 0.0231 | 0.0224 | 0.0245 | 0.0143 | 0.0130 | 0.0141 |
| | | | | | | Linear Probing | | | | | | |
| 0 | **0.4295** | 0.2921 | 0.3671 | **0.6755** | 0.5529 | 0.6109 | **0.6134** | 0.6833 | 0.6460 | **0.4688** | 0.4999 | 0.4773 |
| 1 | **0.4674** | 0.2940 | 0.3585 | **0.6864** | 0.5511 | 0.6060 | **0.6084** | 0.7005 | 0.6677 | **0.4493** | 0.5170 | 0.4878 |
| 2 | **0.4697** | 0.3261 | 0.3420 | **0.6906** | 0.5807 | 0.5873 | **0.5486** | 0.6185 | 0.6111 | **0.4110** | 0.4742 | 0.4615 |
| 3 | **0.3761** | 0.3056 | 0.3723 | **0.6512** | 0.5541 | 0.6117 | **0.6048** | 0.6380 | 0.6066 | **0.4761** | 0.4825 | 0.4524 |
| 4 | **0.4427** | 0.3106 | 0.3821 | **0.6887** | 0.5634 | 0.6216 | **0.5782** | 0.6431 | 0.6088 | **0.4398** | 0.4983 | 0.4645 |
| 5 | **0.4575** | 0.3021 | 0.3726 | **0.6806** | 0.5559 | 0.6159 | **0.5982** | 0.6785 | 0.6433 | **0.4513** | 0.5163 | 0.4889 |
| 6 | **0.4457** | 0.3559 | 0.3660 | **0.6699** | 0.6073 | 0.6080 | **0.5990** | 0.6457 | 0.6406 | **0.4466** | 0.4863 | 0.4764 |
| 7 | **0.4195** | 0.3055 | 0.3083 | **0.6626** | 0.5575 | 0.5595 | **0.5963** | 0.6522 | 0.6509 | **0.4621** | 0.4995 | 0.4872 |
| 8 | **0.4420** | 0.2614 | 0.3545 | **0.6717** | 0.5221 | 0.5964 | **0.5495** | 0.6322 | 0.5910 | **0.4132** | 0.4799 | 0.4474 |
| 9 | **0.4742** | 0.3165 | 0.3734 | **0.6939** | 0.5659 | 0.6175 | **0.5641** | 0.6432 | 0.6158 | **0.4269** | 0.4961 | 0.4597 |
| avg | **0.4424** | 0.3070 | 0.3597 | **0.6771** | 0.5611 | 0.6035 | **0.5860** | 0.6535 | 0.6282 | **0.4445** | 0.4950 | 0.4703 |
| std | 0.0293 | 0.0244 | 0.0213 | 0.0136 | 0.0219 | 0.0185 | 0.0243 | 0.0256 | 0.0246 | 0.0221 | 0.0145 | 0.0152 |

HB: HELM-BERT, MF: MoLFormer–XL, PC: PeptideCLM. Best results per fold are **bolded**, second-best are underlined. ↑: higher is better, ↓: lower is better.

## S2.2 Propedia PPI prediction

Tables S2 and S3 report fold-wise performance for Propedia PPI prediction under Random and aCSM cluster-based splits, respectively. The aCSM split exhibits higher variance across folds

due to the cluster-based partitioning, where each fold tests on distinct protein clusters. Despite this increased variance, the relative ranking of models remains consistent within each fold.

Table S2: Fold-wise performance on Propedia PPI prediction (Random split).

| | ROC-AUC ↑ | | | | PR-AUC ↑ | | | | MCC ↑ | | | | Balanced Acc ↑ | | | |
|---|---|---|---|---|---|---|---|---|---|---|---|---|---|---|---|---|
| Fold | HB | ESM | MF | PC | HB | ESM | MF | PC | HB | ESM | MF | PC | HB | ESM | MF | PC |
| | | | | | | | | MLP | | | | | | | | |
| 0 | 0.9332 | **0.9346** | 0.9205 | 0.9162 | 0.7975 | **0.8124** | 0.7831 | 0.7753 | 0.6407 | **0.6503** | 0.6156 | 0.5993 | 0.8576 | **0.8596** | 0.8434 | 0.8401 |
| 1 | 0.9405 | **0.9430** | 0.9147 | 0.9101 | 0.8264 | **0.8363** | 0.7688 | 0.7606 | 0.6650 | **0.6728** | 0.6142 | 0.5799 | 0.8691 | **0.8740** | 0.8287 | 0.8294 |
| 2 | **0.9472** | 0.9383 | 0.9180 | 0.9315 | **0.8360** | 0.8217 | 0.7821 | 0.8141 | **0.6973** | 0.6452 | 0.5916 | 0.6613 | **0.8783** | 0.8625 | 0.8350 | 0.8548 |
| 3 | **0.9450** | 0.9428 | 0.9338 | 0.9177 | **0.8443** | 0.8411 | 0.8148 | 0.7703 | **0.7036** | 0.6921 | 0.6579 | 0.5840 | **0.8755** | 0.8746 | 0.8641 | 0.8381 |
| 4 | 0.9440 | **0.9492** | 0.9219 | 0.9197 | 0.8352 | **0.8504** | 0.7843 | 0.7833 | 0.6457 | **0.6916** | 0.6099 | 0.5935 | 0.8712 | **0.8808** | 0.8424 | 0.8403 |
| avg | **0.9420** | 0.9416 | 0.9218 | 0.9190 | 0.8279 | **0.8324** | 0.7866 | 0.7807 | **0.6705** | 0.6704 | 0.6178 | 0.6036 | **0.8703** | 0.8703 | 0.8427 | 0.8405 |
| std | 0.0055 | 0.0055 | 0.0073 | 0.0078 | 0.0181 | 0.0152 | 0.0170 | 0.0204 | 0.0289 | 0.0222 | 0.0244 | 0.0331 | 0.0080 | 0.0089 | 0.0134 | 0.0092 |
| | | | | | | | | Linear | | | | | | | | |
| 0 | 0.6111 | **0.6195** | 0.5994 | 0.5964 | 0.2603 | **0.2699** | 0.2612 | 0.2557 | 0.1259 | **0.1289** | 0.1037 | 0.1032 | 0.5779 | **0.5793** | 0.5636 | 0.5641 |
| 1 | 0.6198 | **0.6237** | 0.5970 | 0.6029 | 0.2712 | **0.2729** | 0.2624 | 0.2702 | 0.1316 | **0.1353** | 0.0986 | 0.1148 | 0.5812 | **0.5839** | 0.5605 | 0.5711 |
| 2 | 0.6126 | **0.6206** | 0.5965 | 0.5935 | 0.2693 | **0.2699** | 0.2595 | 0.2534 | 0.1294 | **0.1387** | 0.1047 | 0.1031 | 0.5807 | **0.5847** | 0.5654 | 0.5625 |
| 3 | 0.6114 | **0.6217** | 0.5945 | 0.5908 | **0.2721** | 0.2650 | 0.2595 | 0.2601 | 0.1278 | **0.1315** | 0.0991 | 0.1077 | 0.5790 | **0.5809** | 0.5600 | 0.5673 |
| 4 | 0.6062 | **0.6167** | 0.5870 | 0.5944 | **0.2792** | 0.2754 | 0.2651 | 0.2730 | 0.1121 | **0.1232** | 0.0892 | 0.1044 | 0.5682 | **0.5756** | 0.5552 | 0.5639 |
| avg | 0.6122 | **0.6205** | 0.5949 | 0.5956 | 0.2704 | **0.2706** | 0.2615 | 0.2625 | 0.1254 | **0.1315** | 0.0991 | 0.1066 | 0.5774 | **0.5809** | 0.5609 | 0.5658 |
| std | 0.0049 | 0.0026 | 0.0047 | 0.0045 | 0.0068 | 0.0039 | 0.0023 | 0.0087 | 0.0077 | 0.0059 | 0.0061 | 0.0049 | 0.0053 | 0.0037 | 0.0039 | 0.0035 |

HB: HELM-BERT, ESM: ESM-2 (650M), MF: MoLFormer–XL, PC: PeptideCLM. Best results per fold are **bolded**, second-best are <u>underlined</u>. ↑: higher is better. Results demonstrate consistent performance patterns across all 5 folds.

Table S3: Fold-wise performance on Propedia PPI prediction (aCSM cluster-based split).

| | ROC-AUC ↑ | | | | PR-AUC ↑ | | | | MCC ↑ | | | | Balanced Acc ↑ | | | |
|---|---|---|---|---|---|---|---|---|---|---|---|---|---|---|---|---|
| Fold | HB | ESM | MF | PC | HB | ESM | MF | PC | HB | ESM | MF | PC | HB | ESM | MF | PC |
| | | | | | | | | MLP | | | | | | | | |
| 0 | **0.7722** | 0.7446 | 0.7127 | 0.7212 | **0.5098** | 0.4691 | 0.3995 | 0.4188 | **0.3273** | 0.2696 | 0.2385 | 0.2610 | **0.6912** | 0.6651 | 0.6489 | 0.6572 |
| 1 | **0.8202** | 0.8168 | 0.7768 | 0.7585 | **0.5867** | 0.5703 | 0.5131 | 0.4918 | **0.4057** | 0.3716 | 0.3242 | 0.2964 | **0.7376** | 0.7290 | 0.7006 | 0.6810 |
| 2 | 0.8048 | **0.8275** | 0.7826 | 0.7619 | 0.5539 | **0.5545** | 0.5055 | 0.4852 | 0.3683 | **0.4063** | 0.3433 | 0.3047 | 0.7220 | **0.7412** | 0.7025 | 0.6826 |
| 3 | 0.7351 | **0.7831** | 0.7781 | 0.7615 | 0.4393 | **0.5089** | 0.4767 | 0.4438 | 0.2432 | 0.3011 | **0.3411** | 0.3141 | 0.6501 | 0.6825 | **0.7070** | 0.6960 |
| 4 | 0.7241 | **0.7704** | 0.7078 | 0.7058 | 0.4555 | **0.5285** | 0.4268 | 0.4184 | 0.2417 | **0.3295** | 0.2221 | 0.2227 | 0.6353 | **0.6767** | 0.6383 | 0.6392 |
| avg | 0.7713 | **0.7885** | 0.7516 | 0.7418 | 0.5090 | **0.5263** | 0.4643 | 0.4516 | 0.3172 | **0.3356** | 0.2938 | 0.2798 | 0.6873 | **0.6989** | 0.6795 | 0.6712 |
| std | 0.0420 | 0.0339 | 0.0378 | 0.0264 | 0.0628 | 0.0397 | 0.0496 | 0.0353 | 0.0737 | 0.0545 | 0.0588 | 0.0377 | 0.0443 | 0.0339 | 0.0330 | 0.0227 |
| | | | | | | | | Linear | | | | | | | | |
| 0 | 0.5434 | **0.5471** | 0.5404 | 0.5236 | 0.2154 | **0.2187** | 0.2169 | 0.2083 | **0.0566** | 0.0536 | 0.0528 | 0.0314 | **0.5346** | 0.5314 | 0.5326 | 0.5189 |
| 1 | 0.5610 | **0.5612** | 0.5493 | 0.5492 | **0.2333** | 0.2272 | 0.2249 | 0.2246 | 0.0640 | **0.0669** | 0.0498 | 0.0608 | 0.5395 | **0.5418** | 0.5311 | 0.5373 |
| 2 | 0.5929 | **0.5942** | 0.5701 | 0.5497 | **0.2485** | 0.2352 | 0.2382 | 0.2386 | 0.0882 | 0.0841 | **0.0931** | 0.0405 | 0.5497 | 0.5463 | **0.5573** | 0.5232 |
| 3 | **0.5829** | 0.5719 | 0.5505 | 0.5432 | **0.2357** | 0.2229 | 0.2215 | 0.2155 | **0.0858** | 0.0673 | 0.0616 | 0.0258 | **0.5531** | 0.5366 | 0.5371 | 0.5150 |
| 4 | 0.5478 | **0.5478** | 0.5319 | 0.5420 | 0.2337 | **0.2342** | 0.2172 | 0.2303 | 0.0481 | **0.0516** | 0.0422 | 0.0488 | 0.5301 | **0.5303** | 0.5250 | 0.5305 |
| avg | **0.5656** | 0.5644 | 0.5484 | 0.5415 | **0.2333** | 0.2276 | 0.2237 | 0.2234 | **0.0685** | 0.0647 | 0.0599 | 0.0414 | **0.5414** | 0.5373 | 0.5366 | 0.5250 |
| std | 0.0217 | 0.0195 | 0.0142 | 0.0106 | 0.0118 | 0.0071 | 0.0087 | 0.0119 | 0.0178 | 0.0130 | 0.0198 | 0.0139 | 0.0098 | 0.0068 | 0.0123 | 0.0090 |

HB: HELM-BERT, ESM: ESM-2 (650M), MF: MoLFormer–XL, PC: PeptideCLM. Best results per fold are **bolded**, second-best are <u>underlined</u>. ↑: higher is better. The larger variance across folds reflects heterogeneous test distributions from cluster-based partitioning, where each fold tests on distinct protein clusters.

## S3. Statistical Comparison of Models

We quantitatively compared HELM-BERT against all baseline models using the corrected resampled $t$-test of Nadeau and Bengio, which accounts for dependence between folds in $k$-fold cross-validation. We fixed the significance level at $\alpha = 0.05$ *a priori*. For each task, split, and evaluation protocol, we report the mean and standard deviation of the fold-wise test scores for HELM-BERT and the comparison model, the fold-wise difference $\Delta$ (HELM-BERT − Model), Cohen's $d$ effect size, and the Nadeau–Bengio corrected $p$-value. We summarize the magnitude of effects using Cohen's $d$ (negligible $|d| < 0.2$, small $0.2 \leq |d| < 0.5$, medium $0.5 \leq |d| < 0.8$, large $|d| \geq 0.8$).

## S3.1 CycPeptMPDB permeability prediction

Tables S4, S5, and S6 summarize the statistical comparisons for CycPeptMPDB permeability prediction under full fine-tuning, head fine-tuning, and linear probing, respectively.

Table S4: Corrected resampled $t$-test comparison between HELM-BERT and SMILES-based encoders on CycPeptMPDB permeability prediction (Full Fine-tuning).

| Model | Metric | HELM-BERT | Model | $\Delta$ | $d$ | $p$ | Sig |
|---|---|---|---|---|---|---|---|
| MoLFormer–XL | $R^2$ | $0.7172 \pm 0.0345$ | $0.5776 \pm 0.0434$ | $+0.1396$ | $+3.40$ | $<0.0001$ | *** |
| MoLFormer–XL | Pearson | $0.8493 \pm 0.0207$ | $0.7673 \pm 0.0247$ | $+0.0820$ | $+3.23$ | $<0.0001$ | *** |
| MoLFormer–XL | RMSE | $0.4164 \pm 0.0211$ | $0.5094 \pm 0.0267$ | $-0.0931$ | $-3.33$ | $<0.0001$ | *** |
| MoLFormer–XL | MAE | $0.2946 \pm 0.0102$ | $0.3668 \pm 0.0167$ | $-0.0722$ | $-4.98$ | $<0.0001$ | *** |
| PeptideCLM | $R^2$ | $0.7172 \pm 0.0345$ | $0.5360 \pm 0.0245$ | $+0.1811$ | $+7.18$ | $<0.0001$ | *** |
| PeptideCLM | Pearson | $0.8493 \pm 0.0207$ | $0.7413 \pm 0.0127$ | $+0.1080$ | $+6.04$ | $<0.0001$ | *** |
| PeptideCLM | RMSE | $0.4164 \pm 0.0211$ | $0.5344 \pm 0.0158$ | $-0.1180$ | $-5.67$ | $<0.0001$ | *** |
| PeptideCLM | MAE | $0.2946 \pm 0.0102$ | $0.3847 \pm 0.0109$ | $-0.0901$ | $-8.30$ | $<0.0001$ | *** |

For each metric, we report mean $\pm$ std over folds, fold-wise difference $\Delta$ (HELM-BERT $-$ Model), Cohen's $d$ effect size, and $p$-value (corrected resampled $t$-test with FDR correction). Significance codes: * $p < 0.05$, ** $p < 0.01$, *** $p < 0.001$.

Table S5: Corrected resampled $t$-test comparison between HELM-BERT and SMILES-based encoders on CycPeptMPDB permeability prediction (Head Fine-tuning, frozen encoders).

| Model | Metric | HELM-BERT | Model | $\Delta$ | $d$ | $p$ | Sig |
|---|---|---|---|---|---|---|---|
| MoLFormer–XL | $R^2$ | $0.6181 \pm 0.0343$ | $0.5510 \pm 0.0348$ | $+0.0670$ | $+3.40$ | $<0.0001$ | *** |
| MoLFormer–XL | Pearson | $0.7906 \pm 0.0199$ | $0.7446 \pm 0.0218$ | $+0.0460$ | $+3.47$ | $<0.0001$ | *** |
| MoLFormer–XL | RMSE | $0.4845 \pm 0.0231$ | $0.5255 \pm 0.0224$ | $-0.0409$ | $-3.29$ | $<0.0001$ | *** |
| MoLFormer–XL | MAE | $0.3527 \pm 0.0143$ | $0.3914 \pm 0.0130$ | $-0.0387$ | $-4.21$ | $<0.0001$ | *** |
| PeptideCLM | $R^2$ | $0.6181 \pm 0.0343$ | $0.4297 \pm 0.0256$ | $+0.1883$ | $+4.85$ | $<0.0001$ | *** |
| PeptideCLM | Pearson | $0.7906 \pm 0.0199$ | $0.6569 \pm 0.0193$ | $+0.1337$ | $+5.49$ | $<0.0001$ | *** |
| PeptideCLM | RMSE | $0.4845 \pm 0.0231$ | $0.5927 \pm 0.0245$ | $-0.1082$ | $-4.29$ | $<0.0001$ | *** |
| PeptideCLM | MAE | $0.3527 \pm 0.0143$ | $0.4426 \pm 0.0141$ | $-0.0899$ | $-6.55$ | $<0.0001$ | *** |

Columns are as in Table S4. Significance codes: * $p < 0.05$, ** $p < 0.01$, *** $p < 0.001$.

Table S6: Corrected resampled $t$-test comparison between HELM-BERT and SMILES-based encoders on CycPeptMPDB permeability prediction (Linear Probing, frozen encoders).

| Model | Metric | HELM-BERT | Model | $\Delta$ | $d$ | $p$ | Sig |
|---|---|---|---|---|---|---|---|
| MoLFormer–XL | $R^2$ | $0.4424 \pm 0.0293$ | $0.3070 \pm 0.0244$ | $+0.1355$ | $+3.84$ | $<0.0001$ | *** |
| MoLFormer–XL | Pearson | $0.6771 \pm 0.0136$ | $0.5611 \pm 0.0219$ | $+0.1160$ | $+4.82$ | $<0.0001$ | *** |
| MoLFormer–XL | RMSE | $0.5860 \pm 0.0243$ | $0.6535 \pm 0.0256$ | $-0.0675$ | $-3.77$ | $<0.0001$ | *** |
| MoLFormer–XL | MAE | $0.4445 \pm 0.0221$ | $0.4950 \pm 0.0145$ | $-0.0505$ | $-2.41$ | $0.0007$ | *** |
| PeptideCLM | $R^2$ | $0.4424 \pm 0.0293$ | $0.3597 \pm 0.0213$ | $+0.0827$ | $+2.37$ | $0.0007$ | *** |
| PeptideCLM | Pearson | $0.6771 \pm 0.0136$ | $0.6035 \pm 0.0185$ | $+0.0736$ | $+3.84$ | $<0.0001$ | *** |
| PeptideCLM | RMSE | $0.5860 \pm 0.0243$ | $0.6282 \pm 0.0246$ | $-0.0421$ | $-2.38$ | $0.0007$ | *** |
| PeptideCLM | MAE | $0.4445 \pm 0.0221$ | $0.4703 \pm 0.0152$ | $-0.0258$ | $-1.26$ | $0.0231$ | * |

Columns are as in Table S4. Significance codes: * $p < 0.05$, ** $p < 0.01$, *** $p < 0.001$.

## S3.2 HELM-BERT ablation studies

Tables S7 and S8 report corrected resampled $t$-tests comparing HELM-BERT (full) to architectural ablations and pre-training data ablations, respectively.

Table S7: Corrected resampled $t$-test comparison between HELM-BERT and architectural ablations on CycPeptMPDB permeability prediction (Full Fine-tuning).

| Variant | Metric | HELM-BERT (full) | Variant | $\Delta$ | $d$ | $p$ | Sig |
|---|---|---|---|---|---|---|---|
| w/o Disentangled Attn. | $R^2$ | $0.7172 \pm 0.0345$ | $0.6683 \pm 0.0303$ | $+0.0489$ | $+2.90$ | $0.0010$ | *** |
| w/o Disentangled Attn. | Pearson | $0.8493 \pm 0.0207$ | $0.8223 \pm 0.0184$ | $+0.0270$ | $+2.89$ | $0.0010$ | *** |
| w/o Disentangled Attn. | RMSE | $0.4164 \pm 0.0211$ | $0.4515 \pm 0.0202$ | $-0.0351$ | $-2.54$ | $0.0018$ | ** |
| w/o Disentangled Attn. | MAE | $0.2946 \pm 0.0102$ | $0.3275 \pm 0.0134$ | $-0.0330$ | $-3.38$ | $0.0009$ | *** |
| w/o EMD | $R^2$ | $0.7172 \pm 0.0345$ | $0.7045 \pm 0.0394$ | $+0.0126$ | $+0.66$ | $0.2872$ | |
| w/o EMD | Pearson | $0.8493 \pm 0.0207$ | $0.8413 \pm 0.0224$ | $+0.0079$ | $+0.78$ | $0.2740$ | |
| w/o EMD | RMSE | $0.4164 \pm 0.0211$ | $0.4256 \pm 0.0268$ | $-0.0092$ | $-0.67$ | $0.2872$ | |
| w/o EMD | MAE | $0.2946 \pm 0.0102$ | $0.3013 \pm 0.0152$ | $-0.0067$ | $-0.59$ | $0.2876$ | |
| w/o nGiE | $R^2$ | $0.7172 \pm 0.0345$ | $0.7129 \pm 0.0410$ | $+0.0042$ | $+0.27$ | $0.6004$ | |
| w/o nGiE | Pearson | $0.8493 \pm 0.0207$ | $0.8462 \pm 0.0227$ | $+0.0031$ | $+0.40$ | $0.4801$ | |
| w/o nGiE | RMSE | $0.4164 \pm 0.0211$ | $0.4192 \pm 0.0259$ | $-0.0028$ | $-0.25$ | $0.6004$ | |
| w/o nGiE | MAE | $0.2946 \pm 0.0102$ | $0.2973 \pm 0.0118$ | $-0.0028$ | $-0.32$ | $0.5633$ | |
| w/o Span Masking | $R^2$ | $0.7172 \pm 0.0345$ | $0.7064 \pm 0.0446$ | $+0.0108$ | $+0.63$ | $0.2872$ | |
| w/o Span Masking | Pearson | $0.8493 \pm 0.0207$ | $0.8427 \pm 0.0260$ | $+0.0066$ | $+0.69$ | $0.2872$ | |
| w/o Span Masking | RMSE | $0.4164 \pm 0.0211$ | $0.4239 \pm 0.0290$ | $-0.0075$ | $-0.64$ | $0.2872$ | |
| w/o Span Masking | MAE | $0.2946 \pm 0.0102$ | $0.3007 \pm 0.0172$ | $-0.0062$ | $-0.60$ | $0.2876$ | |
| Vanilla-BERT | $R^2$ | $0.7172 \pm 0.0345$ | $0.6523 \pm 0.0546$ | $+0.0649$ | $+1.65$ | $0.0146$ | * |
| Vanilla-BERT | Pearson | $0.8493 \pm 0.0207$ | $0.8114 \pm 0.0299$ | $+0.0379$ | $+1.75$ | $0.0121$ | * |
| Vanilla-BERT | RMSE | $0.4164 \pm 0.0211$ | $0.4616 \pm 0.0361$ | $-0.0453$ | $-1.79$ | $0.0121$ | * |
| Vanilla-BERT | MAE | $0.2946 \pm 0.0102$ | $0.3289 \pm 0.0236$ | $-0.0344$ | $-1.90$ | $0.0101$ | * |

For each metric, we report mean $\pm$ std over folds, fold-wise difference $\Delta$ (HELM-BERT $-$ Variant), Cohen's $d$ effect size, and $p$-value (corrected resampled $t$-test with FDR correction). Significance codes: * $p < 0.05$, ** $p < 0.01$, *** $p < 0.001$.

Table S8: Corrected resampled $t$-test comparison between HELM-BERT and pre-training data ablations on CycPeptMPDB permeability prediction (Full Fine-tuning).

| Variant | Metric | HELM-BERT (full) | Variant | $\Delta$ | $d$ | $p$ | Sig |
|---|---|---|---|---|---|---|---|
| w/o ChEMBL | $R^2$ | $0.7172 \pm 0.0345$ | $0.7024 \pm 0.0434$ | $+0.0147$ | $+1.08$ | $0.0980$ | |
| w/o ChEMBL | Pearson | $0.8493 \pm 0.0207$ | $0.8403 \pm 0.0246$ | $+0.0090$ | $+1.17$ | $0.0980$ | |
| w/o ChEMBL | RMSE | $0.4164 \pm 0.0211$ | $0.4268 \pm 0.0283$ | $-0.0104$ | $-1.10$ | $0.0980$ | |
| w/o ChEMBL | MAE | $0.2946 \pm 0.0102$ | $0.3031 \pm 0.0192$ | $-0.0085$ | $-0.87$ | $0.1786$ | |
| w/o Propedia | $R^2$ | $0.7172 \pm 0.0345$ | $0.7259 \pm 0.0469$ | $-0.0088$ | $-0.56$ | $0.3643$ | |
| w/o Propedia | Pearson | $0.8493 \pm 0.0207$ | $0.8542 \pm 0.0267$ | $-0.0049$ | $-0.66$ | $0.3335$ | |
| w/o Propedia | RMSE | $0.4164 \pm 0.0211$ | $0.4091 \pm 0.0319$ | $+0.0072$ | $+0.60$ | $0.3563$ | |
| w/o Propedia | MAE | $0.2946 \pm 0.0102$ | $0.2922 \pm 0.0181$ | $+0.0024$ | $+0.26$ | $0.6215$ | |
| w/o CycPeptMPDB | $R^2$ | $0.7172 \pm 0.0345$ | $0.7094 \pm 0.0509$ | $+0.0078$ | $+0.35$ | $0.5836$ | |
| w/o CycPeptMPDB | Pearson | $0.8493 \pm 0.0207$ | $0.8453 \pm 0.0281$ | $+0.0039$ | $+0.34$ | $0.5836$ | |
| w/o CycPeptMPDB | RMSE | $0.4164 \pm 0.0211$ | $0.4213 \pm 0.0340$ | $-0.0049$ | $-0.31$ | $0.5961$ | |
| w/o CycPeptMPDB | MAE | $0.2946 \pm 0.0102$ | $0.2976 \pm 0.0207$ | $-0.0031$ | $-0.23$ | $0.6233$ | |
| From scratch | $R^2$ | $0.7172 \pm 0.0345$ | $0.6644 \pm 0.0411$ | $+0.0528$ | $+2.48$ | $0.0024$ | ** |
| From scratch | Pearson | $0.8493 \pm 0.0207$ | $0.8170 \pm 0.0246$ | $+0.0323$ | $+2.36$ | $0.0024$ | ** |
| From scratch | RMSE | $0.4164 \pm 0.0211$ | $0.4537 \pm 0.0275$ | $-0.0373$ | $-2.42$ | $0.0024$ | ** |
| From scratch | MAE | $0.2946 \pm 0.0102$ | $0.3278 \pm 0.0133$ | $-0.0333$ | $-4.20$ | $0.0001$ | *** |

For each metric, we report mean $\pm$ std over folds, fold-wise difference $\Delta$ (HELM-BERT $-$ Variant), Cohen's $d$ effect size, and $p$-value (corrected resampled $t$-test with FDR correction). Significance codes: * $p < 0.05$, ** $p < 0.01$, *** $p < 0.001$.

## S3.3 Propedia PPI prediction

Tables S9 and S10 summarize the corrected resampled $t$-tests for the Random Split under MLP head fine-tuning and linear probing, respectively. Tables S11 and S12 report the corresponding results for the aCSM Cluster-based Split.

Table S9: Corrected resampled $t$-test comparison between HELM-BERT and baseline models on Propedia PPI prediction (Random Split, MLP head).

| Model | Metric | HELM-BERT | Model | Δ | $d$ | $p$ | Sig |
|---|---|---|---|---|---|---|---|
| ESM-2 (650M) | ROC-AUC | 0.9420 ± 0.0055 | 0.9416 ± 0.0055 | +0.0004 | +0.07 | 0.9985 | |
| ESM-2 (650M) | PR-AUC | 0.8279 ± 0.0181 | 0.8324 ± 0.0152 | −0.0045 | −0.35 | 0.8180 | |
| ESM-2 (650M) | Balanced Acc. | 0.8703 ± 0.0080 | 0.8703 ± 0.0089 | +0.0000 | +0.00 | 0.9985 | |
| ESM-2 (650M) | MCC | 0.6705 ± 0.0289 | 0.6704 ± 0.0222 | +0.0001 | +0.00 | 0.9985 | |
| ESM-2 (150M) | ROC-AUC | 0.9420 ± 0.0055 | 0.9434 ± 0.0031 | −0.0015 | −0.29 | 0.8180 | |
| ESM-2 (150M) | PR-AUC | 0.8279 ± 0.0181 | 0.8418 ± 0.0076 | −0.0139 | −0.83 | 0.5678 | |
| ESM-2 (150M) | Balanced Acc. | 0.8703 ± 0.0080 | 0.8740 ± 0.0037 | −0.0037 | −0.45 | 0.8180 | |
| ESM-2 (150M) | MCC | 0.6705 ± 0.0289 | 0.6822 ± 0.0099 | −0.0117 | −0.32 | 0.8180 | |
| ESM-2 (35M) | ROC-AUC | 0.9420 ± 0.0055 | 0.9466 ± 0.0055 | −0.0046 | −0.62 | 0.6959 | |
| ESM-2 (35M) | PR-AUC | 0.8279 ± 0.0181 | 0.8474 ± 0.0152 | −0.0196 | −1.04 | 0.4289 | |
| ESM-2 (35M) | Balanced Acc. | 0.8703 ± 0.0080 | 0.8787 ± 0.0073 | −0.0084 | −0.71 | 0.6487 | |
| ESM-2 (35M) | MCC | 0.6705 ± 0.0289 | 0.6886 ± 0.0252 | −0.0181 | −0.44 | 0.8180 | |
| PeptideCLM | ROC-AUC | 0.9420 ± 0.0055 | 0.9190 ± 0.0078 | +0.0229 | +3.58 | 0.1084 | |
| PeptideCLM | PR-AUC | 0.8279 ± 0.0181 | 0.7807 ± 0.0204 | +0.0472 | +1.94 | 0.1438 | |
| PeptideCLM | Balanced Acc. | 0.8703 ± 0.0080 | 0.8405 ± 0.0092 | +0.0298 | +3.18 | 0.1084 | |
| PeptideCLM | MCC | 0.6705 ± 0.0289 | 0.6036 ± 0.0331 | +0.0669 | +1.91 | 0.1438 | |
| MoLFormer–XL | ROC-AUC | 0.9420 ± 0.0055 | 0.9218 ± 0.0073 | +0.0202 | +2.53 | 0.1416 | |
| MoLFormer–XL | PR-AUC | 0.8279 ± 0.0181 | 0.7866 ± 0.0170 | +0.0413 | +2.22 | 0.1416 | |
| MoLFormer–XL | Balanced Acc. | 0.8703 ± 0.0080 | 0.8427 ± 0.0134 | +0.0276 | +1.89 | 0.1438 | |
| MoLFormer–XL | MCC | 0.6705 ± 0.0289 | 0.6178 ± 0.0244 | +0.0526 | +1.68 | 0.1766 | |
| Peptide Descriptors | ROC-AUC | 0.9420 ± 0.0055 | 0.9320 ± 0.0038 | +0.0100 | +2.31 | 0.1416 | |
| Peptide Descriptors | PR-AUC | 0.8279 ± 0.0181 | 0.8233 ± 0.0090 | +0.0045 | +0.26 | 0.8180 | |
| Peptide Descriptors | Balanced Acc. | 0.8703 ± 0.0080 | 0.8619 ± 0.0061 | +0.0084 | +1.43 | 0.2413 | |
| Peptide Descriptors | MCC | 0.6705 ± 0.0289 | 0.6589 ± 0.0144 | +0.0116 | +0.36 | 0.8180 | |

For each metric, we report mean ± std over folds, fold-wise difference Δ (HELM-BERT − Model), Cohen's $d$ effect size, and $p$-value (corrected resampled $t$-test with FDR correction). Significance codes: * $p < 0.05$, ** $p < 0.01$, *** $p < 0.001$.

Table S10: Corrected resampled *t*-test comparison between HELM-BERT and baseline models on Propedia PPI prediction (Random Split, Linear probing).

| Model | Metric | HELM-BERT | Model | Δ | d | p | Sig |
|---|---|---|---|---|---|---|---|
| ESM-2 (650M) | ROC-AUC | $0.6122 \pm 0.0049$ | $0.6205 \pm 0.0026$ | −0.0083 | −3.13 | 0.0255 | * |
| ESM-2 (650M) | PR-AUC | $0.2704 \pm 0.0068$ | $0.2706 \pm 0.0039$ | −0.0002 | −0.02 | 0.9727 | |
| ESM-2 (650M) | Balanced Acc. | $0.5774 \pm 0.0053$ | $0.5809 \pm 0.0037$ | −0.0035 | −1.44 | 0.1463 | |
| ESM-2 (650M) | MCC | $0.1254 \pm 0.0077$ | $0.1315 \pm 0.0059$ | −0.0062 | −1.65 | 0.1195 | |
| ESM-2 (150M) | ROC-AUC | $0.6122 \pm 0.0049$ | $0.6109 \pm 0.0031$ | +0.0013 | +0.32 | 0.6861 | |
| ESM-2 (150M) | PR-AUC | $0.2704 \pm 0.0068$ | $0.2626 \pm 0.0042$ | +0.0079 | +1.97 | 0.0790 | |
| ESM-2 (150M) | Balanced Acc. | $0.5774 \pm 0.0053$ | $0.5757 \pm 0.0026$ | +0.0017 | +0.35 | 0.6861 | |
| ESM-2 (150M) | MCC | $0.1254 \pm 0.0077$ | $0.1235 \pm 0.0039$ | +0.0019 | +0.35 | 0.6861 | |
| ESM-2 (35M) | ROC-AUC | $0.6122 \pm 0.0049$ | $0.6047 \pm 0.0041$ | +0.0075 | +1.26 | 0.1688 | |
| ESM-2 (35M) | PR-AUC | $0.2704 \pm 0.0068$ | $0.2557 \pm 0.0053$ | +0.0148 | +2.70 | 0.0366 | * |
| ESM-2 (35M) | Balanced Acc. | $0.5774 \pm 0.0053$ | $0.5719 \pm 0.0038$ | +0.0054 | +1.41 | 0.1463 | |
| ESM-2 (35M) | MCC | $0.1254 \pm 0.0077$ | $0.1176 \pm 0.0066$ | +0.0078 | +1.23 | 0.1688 | |
| PeptideCLM | ROC-AUC | $0.6122 \pm 0.0049$ | $0.5956 \pm 0.0045$ | +0.0166 | +4.73 | 0.0073 | ** |
| PeptideCLM | PR-AUC | $0.2704 \pm 0.0068$ | $0.2625 \pm 0.0087$ | +0.0080 | +1.34 | 0.1562 | |
| PeptideCLM | Balanced Acc. | $0.5774 \pm 0.0053$ | $0.5658 \pm 0.0035$ | +0.0116 | +2.26 | 0.0558 | |
| PeptideCLM | MCC | $0.1254 \pm 0.0077$ | $0.1066 \pm 0.0049$ | +0.0187 | +2.65 | 0.0366 | * |
| MoLFormer–XL | ROC-AUC | $0.6122 \pm 0.0049$ | $0.5949 \pm 0.0047$ | +0.0174 | +4.26 | 0.0094 | ** |
| MoLFormer–XL | PR-AUC | $0.2704 \pm 0.0068$ | $0.2615 \pm 0.0023$ | +0.0089 | +1.52 | 0.1383 | |
| MoLFormer–XL | Balanced Acc. | $0.5774 \pm 0.0053$ | $0.5609 \pm 0.0039$ | +0.0165 | +5.03 | 0.0068 | ** |
| MoLFormer–XL | MCC | $0.1254 \pm 0.0077$ | $0.0991 \pm 0.0061$ | +0.0263 | +5.85 | 0.0057 | ** |
| Peptide Descriptors | ROC-AUC | $0.6122 \pm 0.0049$ | $0.5584 \pm 0.0041$ | +0.0538 | +12.58 | 0.0011 | ** |
| Peptide Descriptors | PR-AUC | $0.2704 \pm 0.0068$ | $0.2311 \pm 0.0039$ | +0.0393 | +5.35 | 0.0064 | ** |
| Peptide Descriptors | Balanced Acc. | $0.5774 \pm 0.0053$ | $0.5412 \pm 0.0081$ | +0.0362 | +5.89 | 0.0057 | ** |
| Peptide Descriptors | MCC | $0.1254 \pm 0.0077$ | $0.0665 \pm 0.0127$ | +0.0588 | +6.14 | 0.0057 | ** |

For each metric, we report mean ± std over folds, fold-wise difference Δ (HELM-BERT − Model), Cohen's *d* effect size, and *p*-value (corrected resampled *t*-test with FDR correction). Significance codes: * $p < 0.05$, ** $p < 0.01$, *** $p < 0.001$.

Table S11: Corrected resampled *t*-test comparison between HELM-BERT and baseline models on Propedia PPI prediction (aCSM Cluster Split, MLP head).

| Model | Metric | HELM-BERT | Model | Δ | d | p | Sig |
|---|---|---|---|---|---|---|---|
| ESM-2 (650M) | ROC-AUC | $0.7713 \pm 0.0420$ | $0.7885 \pm 0.0339$ | −0.0172 | −0.53 | 0.8997 | |
| ESM-2 (650M) | PR-AUC | $0.5090 \pm 0.0628$ | $0.5263 \pm 0.0397$ | −0.0173 | −0.34 | 0.8997 | |
| ESM-2 (650M) | Balanced Acc. | $0.6873 \pm 0.0443$ | $0.6989 \pm 0.0339$ | −0.0116 | −0.41 | 0.8997 | |
| ESM-2 (650M) | MCC | $0.3172 \pm 0.0737$ | $0.3356 \pm 0.0545$ | −0.0184 | −0.30 | 0.8997 | |
| ESM-2 (150M) | ROC-AUC | $0.7713 \pm 0.0420$ | $0.7789 \pm 0.0473$ | −0.0076 | −0.12 | 0.9056 | |
| ESM-2 (150M) | PR-AUC | $0.5090 \pm 0.0628$ | $0.5118 \pm 0.0718$ | −0.0027 | −0.03 | 0.9715 | |
| ESM-2 (150M) | Balanced Acc. | $0.6873 \pm 0.0443$ | $0.6971 \pm 0.0426$ | −0.0098 | −0.16 | 0.8997 | |
| ESM-2 (150M) | MCC | $0.3172 \pm 0.0737$ | $0.3367 \pm 0.0775$ | −0.0194 | −0.17 | 0.8997 | |
| ESM-2 (35M) | ROC-AUC | $0.7713 \pm 0.0420$ | $0.7882 \pm 0.0424$ | −0.0169 | −0.35 | 0.8997 | |
| ESM-2 (35M) | PR-AUC | $0.5090 \pm 0.0628$ | $0.5282 \pm 0.0561$ | −0.0192 | −0.26 | 0.8997 | |
| ESM-2 (35M) | Balanced Acc. | $0.6873 \pm 0.0443$ | $0.7046 \pm 0.0330$ | −0.0174 | −0.41 | 0.8997 | |
| ESM-2 (35M) | MCC | $0.3172 \pm 0.0737$ | $0.3464 \pm 0.0669$ | −0.0292 | −0.33 | 0.8997 | |
| PeptideCLM | ROC-AUC | $0.7713 \pm 0.0420$ | $0.7418 \pm 0.0264$ | +0.0295 | +0.84 | 0.8997 | |
| PeptideCLM | PR-AUC | $0.5090 \pm 0.0628$ | $0.4516 \pm 0.0353$ | +0.0574 | +1.38 | 0.8997 | |
| PeptideCLM | Balanced Acc. | $0.6873 \pm 0.0443$ | $0.6712 \pm 0.0227$ | +0.0161 | +0.39 | 0.8997 | |
| PeptideCLM | MCC | $0.3172 \pm 0.0737$ | $0.2798 \pm 0.0377$ | +0.0375 | +0.55 | 0.8997 | |
| MoLFormer–XL | ROC-AUC | $0.7713 \pm 0.0420$ | $0.7516 \pm 0.0378$ | +0.0197 | +0.50 | 0.8997 | |
| MoLFormer–XL | PR-AUC | $0.5090 \pm 0.0628$ | $0.4643 \pm 0.0496$ | +0.0447 | +0.81 | 0.8997 | |
| MoLFormer–XL | Balanced Acc. | $0.6873 \pm 0.0443$ | $0.6795 \pm 0.0330$ | +0.0078 | +0.19 | 0.8997 | |
| MoLFormer–XL | MCC | $0.3172 \pm 0.0737$ | $0.2938 \pm 0.0588$ | +0.0234 | +0.31 | 0.8997 | |
| Peptide Descriptors | ROC-AUC | $0.7713 \pm 0.0420$ | $0.7389 \pm 0.0572$ | +0.0323 | +0.60 | 0.8997 | |
| Peptide Descriptors | PR-AUC | $0.5090 \pm 0.0628$ | $0.4627 \pm 0.0763$ | +0.0463 | +0.63 | 0.8997 | |
| Peptide Descriptors | Balanced Acc. | $0.6873 \pm 0.0443$ | $0.6700 \pm 0.0471$ | +0.0172 | +0.37 | 0.8997 | |
| Peptide Descriptors | MCC | $0.3172 \pm 0.0737$ | $0.2863 \pm 0.0762$ | +0.0309 | +0.41 | 0.8997 | |

Columns are as in Table S9. Significance codes: * $p < 0.05$, ** $p < 0.01$, *** $p < 0.001$.

Table S12: Corrected resampled $t$-test comparison between HELM-BERT and baseline models on Propedia PPI prediction (aCSM Cluster Split, Linear probing).

| Model | Metric | HELM-BERT | Model | $\Delta$ | $d$ | $p$ | Sig |
|---|---|---|---|---|---|---|---|
| ESM-2 (650M) | ROC-AUC | $0.5656 \pm 0.0217$ | $0.5644 \pm 0.0195$ | +0.0012 | +0.21 | 0.7738 | |
| ESM-2 (650M) | PR-AUC | $0.2333 \pm 0.0118$ | $0.2276 \pm 0.0071$ | +0.0057 | +0.75 | 0.4205 | |
| ESM-2 (650M) | Balanced Acc. | $0.5414 \pm 0.0098$ | $0.5373 \pm 0.0068$ | +0.0041 | +0.56 | 0.5003 | |
| ESM-2 (650M) | MCC | $0.0685 \pm 0.0178$ | $0.0647 \pm 0.0130$ | +0.0038 | +0.43 | 0.5794 | |
| ESM-2 (150M) | ROC-AUC | $0.5656 \pm 0.0217$ | $0.5587 \pm 0.0160$ | +0.0069 | +0.74 | 0.4205 | |
| ESM-2 (150M) | PR-AUC | $0.2333 \pm 0.0118$ | $0.2259 \pm 0.0107$ | +0.0075 | +1.04 | 0.3533 | |
| ESM-2 (150M) | Balanced Acc. | $0.5414 \pm 0.0098$ | $0.5354 \pm 0.0086$ | +0.0060 | +1.03 | 0.3533 | |
| ESM-2 (150M) | MCC | $0.0685 \pm 0.0178$ | $0.0590 \pm 0.0161$ | +0.0095 | +1.18 | 0.3533 | |
| ESM-2 (35M) | ROC-AUC | $0.5656 \pm 0.0217$ | $0.5517 \pm 0.0123$ | +0.0139 | +0.89 | 0.4071 | |
| ESM-2 (35M) | PR-AUC | $0.2333 \pm 0.0118$ | $0.2227 \pm 0.0099$ | +0.0106 | +1.44 | 0.3533 | |
| ESM-2 (35M) | Balanced Acc. | $0.5414 \pm 0.0098$ | $0.5317 \pm 0.0148$ | +0.0097 | +0.65 | 0.4678 | |
| ESM-2 (35M) | MCC | $0.0685 \pm 0.0178$ | $0.0520 \pm 0.0232$ | +0.0165 | +0.75 | 0.4205 | |
| PeptideCLM | ROC-AUC | $0.5656 \pm 0.0217$ | $0.5415 \pm 0.0106$ | +0.0241 | +1.44 | 0.3533 | |
| PeptideCLM | PR-AUC | $0.2333 \pm 0.0118$ | $0.2234 \pm 0.0119$ | +0.0099 | +1.57 | 0.3533 | |
| PeptideCLM | Balanced Acc. | $0.5414 \pm 0.0098$ | $0.5250 \pm 0.0090$ | +0.0164 | +1.01 | 0.3533 | |
| PeptideCLM | MCC | $0.0685 \pm 0.0178$ | $0.0414 \pm 0.0139$ | +0.0271 | +1.01 | 0.3533 | |
| MoLFormer–XL | ROC-AUC | $0.5656 \pm 0.0217$ | $0.5484 \pm 0.0142$ | +0.0172 | +1.54 | 0.3533 | |
| MoLFormer–XL | PR-AUC | $0.2333 \pm 0.0118$ | $0.2237 \pm 0.0087$ | +0.0096 | +1.37 | 0.3533 | |
| MoLFormer–XL | Balanced Acc. | $0.5414 \pm 0.0098$ | $0.5366 \pm 0.0123$ | +0.0048 | +0.55 | 0.5003 | |
| MoLFormer–XL | MCC | $0.0685 \pm 0.0178$ | $0.0599 \pm 0.0198$ | +0.0086 | +0.78 | 0.4205 | |
| Peptide Descriptors | ROC-AUC | $0.5656 \pm 0.0217$ | $0.5310 \pm 0.0201$ | +0.0347 | +1.36 | 0.3533 | |
| Peptide Descriptors | PR-AUC | $0.2333 \pm 0.0118$ | $0.2117 \pm 0.0093$ | +0.0216 | +1.35 | 0.3533 | |
| Peptide Descriptors | Balanced Acc. | $0.5414 \pm 0.0098$ | $0.5170 \pm 0.0210$ | +0.0244 | +1.23 | 0.3533 | |
| Peptide Descriptors | MCC | $0.0685 \pm 0.0178$ | $0.0274 \pm 0.0344$ | +0.0411 | +1.16 | 0.3533 | |

Columns are as in Table S9. Significance codes: * $p < 0.05$, ** $p < 0.01$, *** $p < 0.001$.

## S3.4 Embedding quality analysis

Table S13 reports corrected resampled $t$-tests comparing HELM-BERT to SMILES-based encoders on embedding quality metrics using full-dimensional (768-d) representations.

Table S13: Statistical comparison between HELM-BERT and SMILES-based encoders on embedding quality metrics (corresponding to Table 6 in main text; full-dimensional representations).

| Task | Metric | Model | HELM-BERT | Model | $\Delta$ | $d$ | $p$ | Sig |
|---|---|---|---|---|---|---|---|---|
| | | **Physicochemical Properties (Regression)** | | | | | | |
| LogP | $R^2$ | MoLFormer–XL | $0.9535 \pm 0.0049$ | $0.9638 \pm 0.0031$ | $-0.0103$ | $-4.52$ | $0.0030$ | ** |
| | | PeptideCLM | $0.9535 \pm 0.0049$ | $0.9527 \pm 0.0018$ | $+0.0008$ | $+0.17$ | $0.8140$ | |
| Molecular Weight | $R^2$ | MoLFormer–XL | $0.9770 \pm 0.0003$ | $0.9842 \pm 0.0009$ | $-0.0072$ | $-10.83$ | $<0.0001$ | *** |
| | | PeptideCLM | $0.9770 \pm 0.0003$ | $0.9900 \pm 0.0001$ | $-0.0130$ | $-33.92$ | $<0.0001$ | *** |
| TPSA | $R^2$ | MoLFormer–XL | $0.9779 \pm 0.0005$ | $0.9840 \pm 0.0008$ | $-0.0061$ | $-9.65$ | $<0.0001$ | *** |
| | | PeptideCLM | $0.9779 \pm 0.0005$ | $0.9880 \pm 0.0002$ | $-0.0101$ | $-14.79$ | $<0.0001$ | *** |
| | | **Structural Features (Classification)** | | | | | | |
| Structure Type | Accuracy (Linear) | MoLFormer–XL | $0.9996 \pm 0.0002$ | $0.9810 \pm 0.0015$ | $+0.0186$ | $+14.29$ | $<0.0001$ | *** |
| | | PeptideCLM | $0.9996 \pm 0.0002$ | $0.9763 \pm 0.0018$ | $+0.0233$ | $+12.30$ | $<0.0001$ | *** |
| Number of Rings | Accuracy (Linear) | MoLFormer–XL | $0.9975 \pm 0.0006$ | $0.9788 \pm 0.0016$ | $+0.0187$ | $+12.88$ | $<0.0001$ | *** |
| | | PeptideCLM | $0.9975 \pm 0.0006$ | $0.9739 \pm 0.0027$ | $+0.0236$ | $+10.04$ | $<0.0001$ | *** |

For each metric, we report mean $\pm$ std over folds, fold-wise difference $\Delta$ (HELM-BERT $-$ Model), Cohen's $d$ effect size, and $p$-value (corrected resampled $t$-test with FDR correction). Significance codes: * $p < 0.05$, ** $p < 0.01$, *** $p < 0.001$. Statistical tests were applied only to cross-validated metrics.

# S4. Component Activation Analysis

## Methods

For each architectural ablation variant, we loaded the corresponding pre-trained MLM checkpoint and computed simple summary statistics over the encoder weights to characterize the effective activation of individual components:

- **nGiE Norm**: L2 norm of the convolutional kernel in the n-gram Induced Encoder (nGiE).

- **Pos Norm**: L2 norm of the absolute position embedding matrix.

- **Per-Pos Std**: Standard deviation of the per-position L2 norms, reflecting how differently individual positions are encoded.

- **L/C, R/C**: Neighbor-to-center kernel weight ratios, defined as the mean absolute weight of the left/right kernel positions divided by that of the center position.

- **LN Mean**: Mean of the LayerNorm weights applied to the nGiE output, indicating the overall scaling of this component.

These statistics are descriptive summaries of the learned weights in our training setup and are used to compare activation patterns across ablations.

## Results

Table S14: Component activation analysis for ablation variants.

| Variant | nGiE Norm | Pos Norm | Per-Pos Std | L/C | R/C | LN Mean |
| --- | ---: | ---: | ---: | ---: | ---: | ---: |
| HELM-BERT (full) | 32.6 | 21.1 | 0.407 | 1.073 | 1.068 | 0.915 |
| w/o Disentangled Attention | 37.6 | 44.7 | 1.278 | 1.380 | 1.402 | 0.850 |
| w/o nGiE | – | 22.1 | 0.449 | – | – | – |
| w/o Span Masking | 29.2 | 18.2 | 0.280 | 1.059 | 1.059 | 0.948 |
| Vanilla-BERT | – | 15.5 | 0.154 | – | – | – |

nGiE Norm: L2 norm of n-gram encoding layer. Pos Norm: L2 norm of position embeddings. Per-Pos Std: standard deviation across positions. L/C and R/C: ratios of left/right kernel importance to center. LN Mean: LayerNorm weight mean for nGiE output.

The w/o Disentangled Attention variant exhibits substantially different activation patterns compared to all other variants (Table S14). Its nGiE norm is 15% higher and its position embedding norm is more than doubled. Per-position variation is also dramatically higher, which is consistent with a much stronger reliance on absolute position information when disentangled attention is removed in this setting.

## S5. Embedding Visualization

Figure S2 shows PCA projections colored by structure type, complementing the t-SNE visualization in the main text. Figure S3 shows two-dimensional PCA and t-SNE projections of pre-trained embeddings, color-coded by source dataset (CycPeptMPDB v1.2, ChEMBL v35, Propedia v2.3). Figures S4–S6 show the same projections color-coded by physicochemical properties (LogP, exact molecular weight, TPSA). Figure S7 shows projections colored by number of rings.

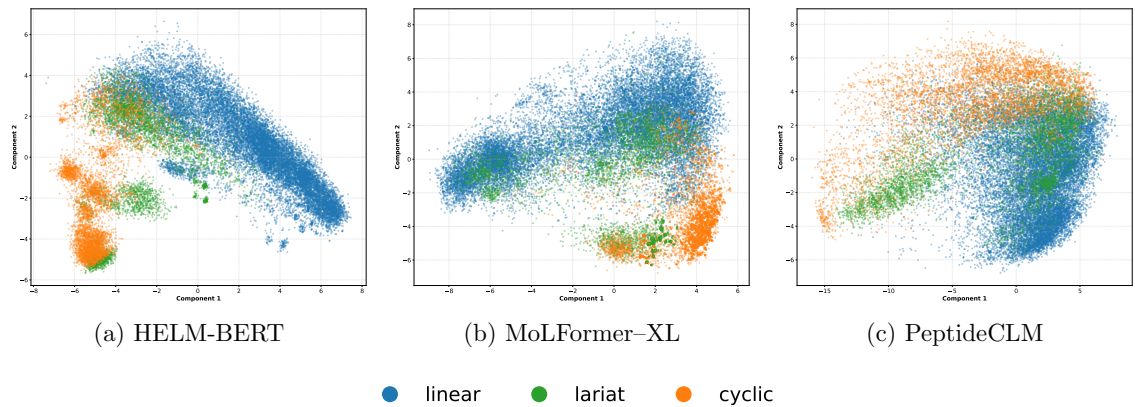

Figure S2: PCA projections of pre-trained embeddings colored by structure type (linear, lariat, cyclic). HELM-BERT shows clearer separation between structural categories compared to SMILES-based encoders.

12

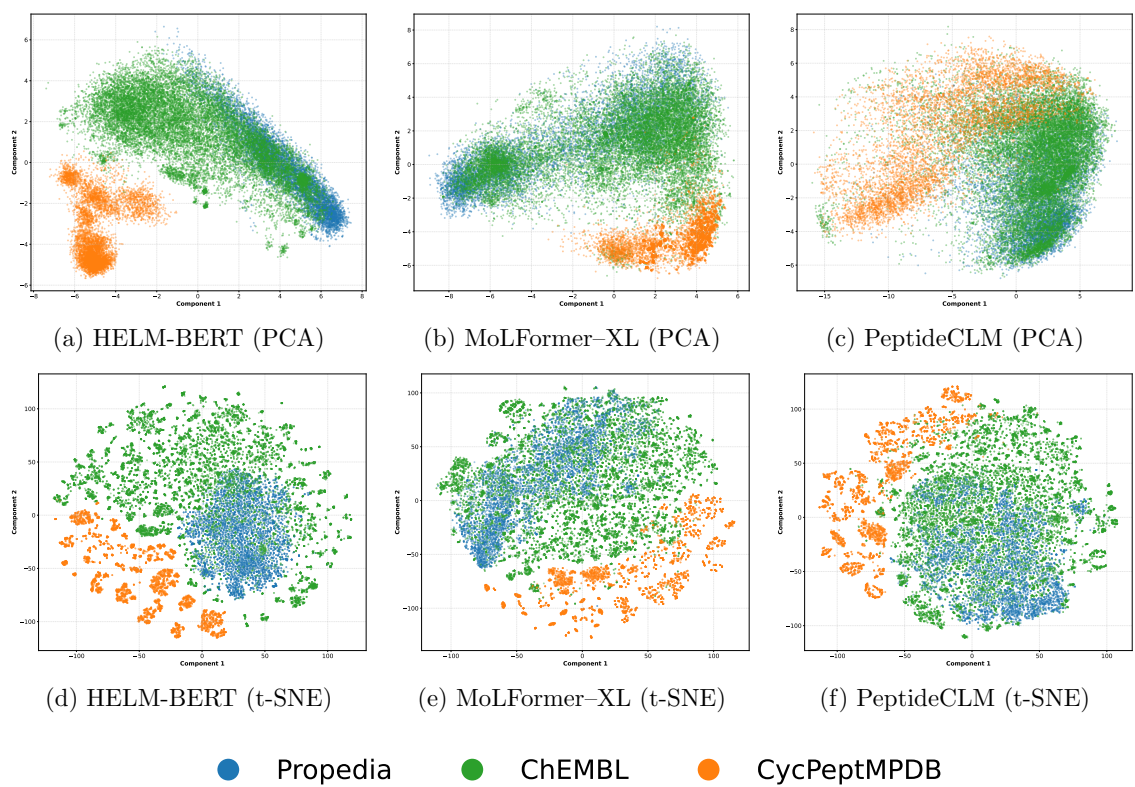

Figure S3: Embedding projections of HELM-BERT, MoLFormer–XL, and PeptideCLM. Top: PCA projection. Bottom: t-SNE projection.

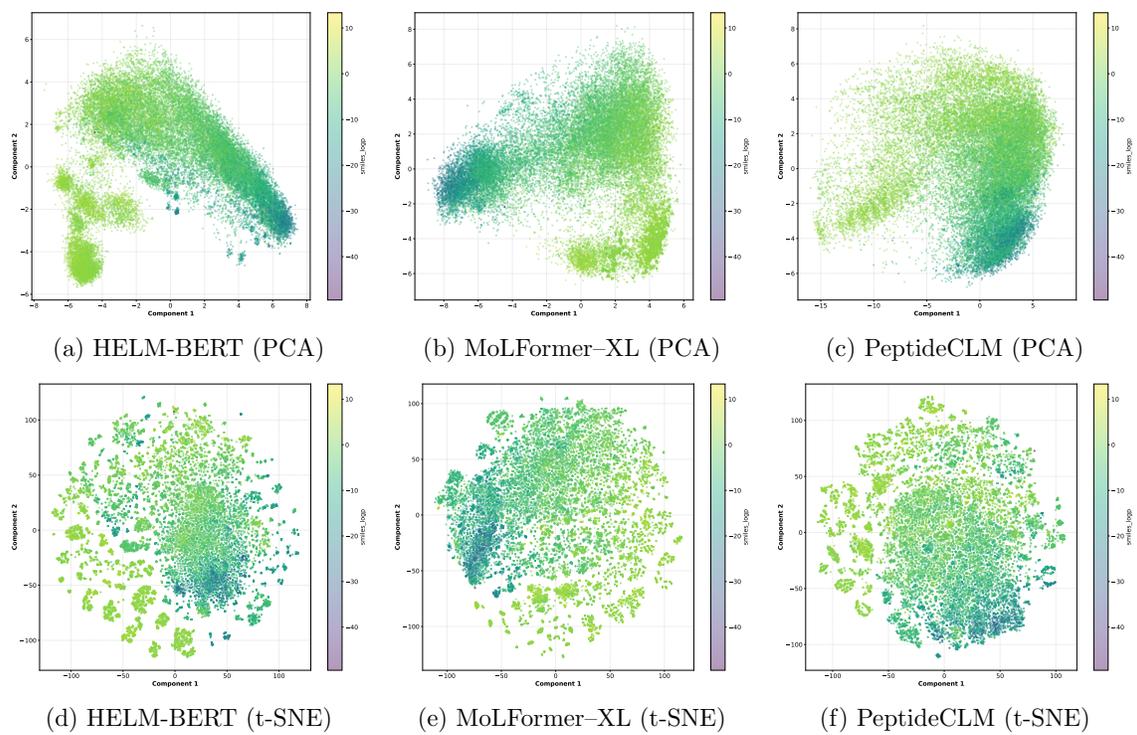

Figure S4: Embedding organization color-coded by LogP. Top: PCA projection. Bottom: t-SNE projection.

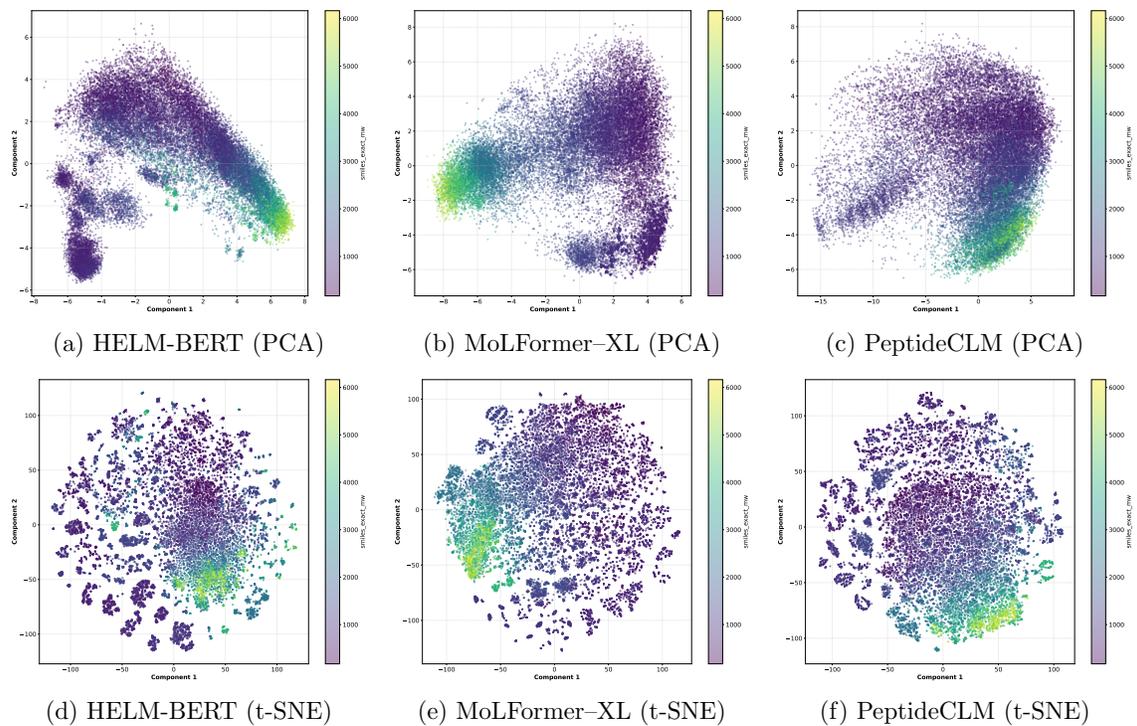

Figure S5: Embedding organization color-coded by exact molecular weight (MW). Top: PCA projection. Bottom: t-SNE projection.

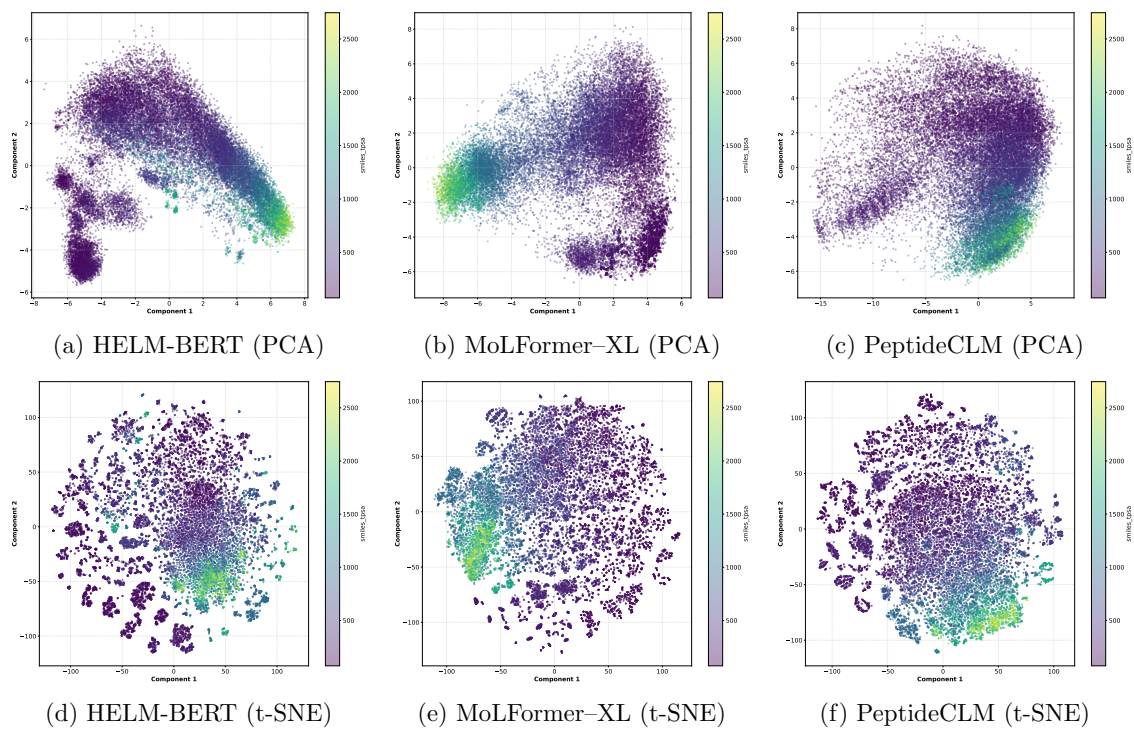

Figure S6: Embedding organization color-coded by TPSA. Top: PCA projection. Bottom: t-SNE projection.

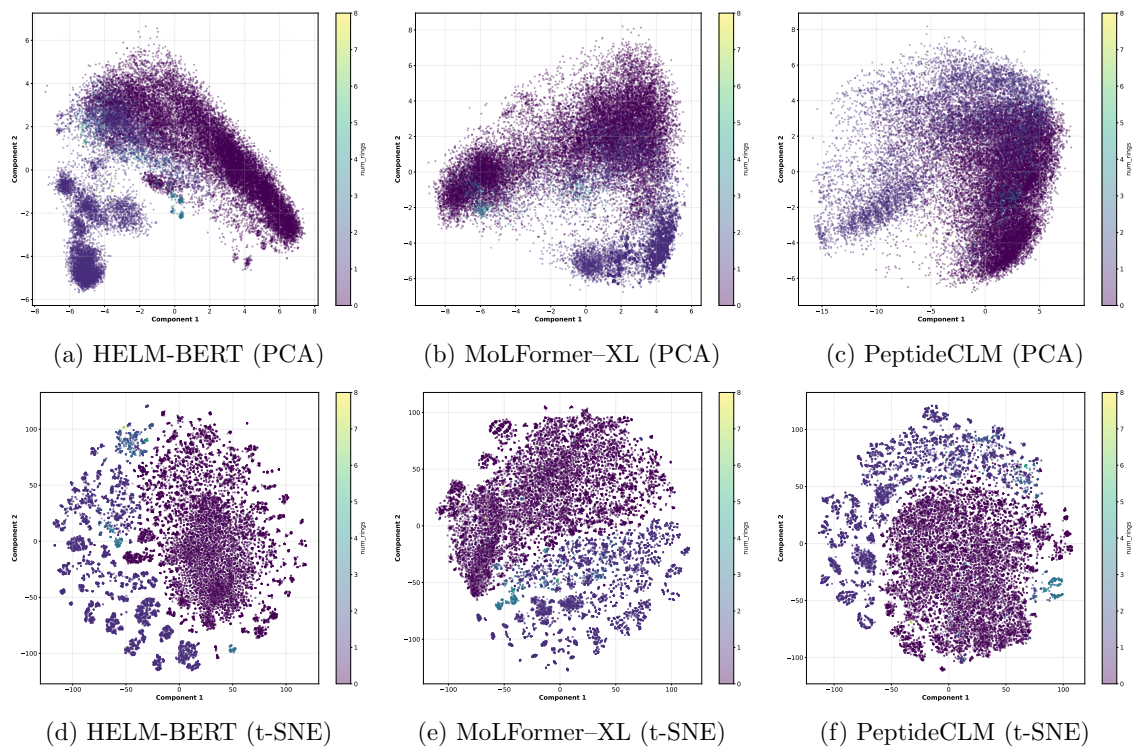

Figure S7: Embedding organization colored by number of rings. Top: PCA projection. Bottom: t-SNE projection.

16

# S6. Low-dimensional Embedding Analysis

Table S15 reports embedding quality metrics computed on 2D PCA projections, complementing the full-dimensional analysis in the main text. While the main text evaluates representations in their original 768-dimensional space, this analysis characterizes the dominant variance structure captured by the first two principal components.

All pairwise comparisons between HELM-BERT and baselines reached statistical significance (corrected resampled $t$-test with FDR correction, $p \leq 0.002$) for structural classification tasks, with large effect sizes ($|d| > 5$). For physicochemical regression, HELM-BERT significantly underperformed MoLFormer–XL on all properties ($p < 0.001$), while differences between HELM-BERT and PeptideCLM were not statistically significant for molecular weight ($p = 0.08$) and TPSA ($p = 0.17$).

Table S15: Embedding quality evaluation of MLM-pre-trained encoders using 2D PCA-reduced representations.

| Task | Metric | HELM-BERT | MoLFormer–XL | PeptideCLM |
|---|---|---|---|---|
| **Physicochemical Properties (Regression)** | | | | |
| LogP | $R^2 \uparrow$ | $0.5956 \pm 0.0043$ | $\mathbf{0.7482 \pm 0.0068}^{\dagger}$ | $0.5513 \pm 0.0068^{\dagger}$ |
|  | MAE $\downarrow$ | $2.95 \pm 0.02$ | $\mathbf{2.29 \pm 0.04}$ | $3.28 \pm 0.04$ |
| Molecular Weight | $R^2 \uparrow$ | $0.5533 \pm 0.0064$ | $\mathbf{0.8225 \pm 0.0030}^{\dagger}$ | $0.5606 \pm 0.0047$ |
|  | MAE $\downarrow$ | $558.79 \pm 4.68$ | $\mathbf{325.86 \pm 5.54}$ | $541.38 \pm 5.62$ |
| TPSA | $R^2 \uparrow$ | $0.6043 \pm 0.0054$ | $\mathbf{0.8372 \pm 0.0030}^{\dagger}$ | $0.6101 \pm 0.0053$ |
|  | MAE $\downarrow$ | $231.31 \pm 1.33$ | $\mathbf{136.76 \pm 2.06}$ | $225.62 \pm 2.62$ |
| **Structural Features (Classification & Separability)** | | | | |
| Structure Type | Accuracy (K-NN) $\uparrow$ | $\mathbf{0.9232}$ | $0.8989$ | $0.8521$ |
|  | Accuracy (Linear) $\uparrow$ | $\mathbf{0.8224 \pm 0.0028}$ | $0.7977 \pm 0.0031^{\dagger}$ | $0.7595 \pm 0.0051^{\dagger}$ |
|  | MCC (Linear) $\uparrow$ | $\mathbf{0.6495}$ | $0.5917$ | $0.4989$ |
|  | Silhouette $\uparrow$ | $\mathbf{0.2309}$ | $0.1321$ | $0.1735$ |
|  | Davies-Bouldin $\downarrow$ | $\mathbf{1.9639}$ | $2.0606$ | $3.0573$ |
|  | Calinski-Harabasz $\uparrow$ | $\mathbf{12900}$ | $6438$ | $4459$ |
| Number of Rings | Accuracy (K-NN) $\uparrow$ | $\mathbf{0.9411}$ | $0.9076$ | $0.8677$ |
|  | Accuracy (Linear) $\uparrow$ | $\mathbf{0.8797 \pm 0.0029}$ | $0.8535 \pm 0.0037^{\dagger}$ | $0.8034 \pm 0.0037^{\dagger}$ |
|  | MCC (Linear) $\uparrow$ | $\mathbf{0.7403}$ | $0.6802$ | $0.5699$ |
|  | Silhouette $\uparrow$ | $\mathbf{-0.0831}$ | $-0.2471$ | $-0.3519$ |
|  | Davies-Bouldin $\downarrow$ | $\mathbf{2.6174}$ | $4.2352$ | $4.3682$ |
|  | Calinski-Harabasz $\uparrow$ | $\mathbf{3230}$ | $1579$ | $1219$ |

Linear probing and K-NN classification assess predictive performance; cluster validity indices (applied to ground-truth labels) quantify class separability in 2D PCA space. Best results are **bolded**, second-best are underlined. $\uparrow$: higher is better, $\downarrow$: lower is better. $^{\dagger}$ indicates significant difference from HELM-BERT (corrected resampled $t$-test with FDR correction, $p < 0.05$); statistical tests were applied only to cross-validated metrics ($R^2$ and Linear Accuracy). For structural classification, all comparisons reached significance ($p \leq 0.0012$, $|d| > 5$). For regression, MoLFormer–XL significantly outperformed HELM-BERT on all properties ($p < 0.0001$); PeptideCLM differed significantly only for LogP ($p = 0.0005$, $d = 7.51$), not for MW ($p = 0.0798$) or TPSA ($p = 0.1733$).



\end{document}